\documentclass{article}
\def\PREPRINT{}  

\usepackage{iclr2027_conference,times}

\ifdefined\PREPRINT
  \iclrfinalcopy
  \newcommand{\pdfauthorname}{Parsa Mazaheri}
\else
  \newcommand{\pdfauthorname}{Anonymous}
\fi

\usepackage{amsmath,amsfonts,bm}

\def\eqref#1{equation~\ref{#1}}

\def\1{\bm{1}}

\DeclareMathAlphabet{\mathsfit}{\encodingdefault}{\sfdefault}{m}{sl}
\SetMathAlphabet{\mathsfit}{bold}{\encodingdefault}{\sfdefault}{bx}{n}

\usepackage{hyperref}
\hypersetup{
  pdftitle={Gathered, Not Admitted: How Attention Brings a Latent Variable into
Verbalizable Form},
  pdfauthor={\pdfauthorname},
  pdfsubject={Interpretability}
}
\usepackage{url}
\usepackage{booktabs}
\usepackage{multirow}
\usepackage{graphicx}
\usepackage{amsmath}
\usepackage{xcolor}

\newcommand{\Rz}{R_{z}}
\newcommand{\Lz}{L_{z}}
\newcommand{\Mz}{M_{z}}
\newcommand{\ci}[3]{#1\,[#2, #3]}

\title{Gathered, Not Admitted: How Attention Brings a Latent Variable into Verbalizable Form}

\author{Parsa Mazaheri \\
University of California, Santa Cruz \\
\texttt{pmazaher@ucsc.edu}}

\begin{document}

\maketitle
\ifdefined\PREPRINT
  \lhead{}%
  \begingroup
  \renewcommand{\thefootnote}{}%
  \footnotetext{Code: \url{https://github.com/parsa-mz/innerj}}%
  \endgroup
\fi

\begin{abstract}
Language models hold latent quantities in a form they can report on, and more of a quantity is
present in that form when the task requires reusing it flexibly. What causes a representation to
enter that form is open, and the word \emph{workspace} invites an admission story: a gate that
decides what gets in. Testing it on open-weight models with Jacobian lenses, over a
benchmark whose five arms share an identical context, we find no gate where it predicts one. Demand
raises a concept's lens visibility beyond what applying an operator to a \emph{supplied} value
produces: $\ci{+0.050}{+0.045}{+0.057}$ in percentile rank on our primary checkpoint, positive on
all four we measure, though that arm answers at ceiling and the accuracy-matched contrast is
stronger under that readout. At the same time one shared linear map decodes the variable from every arm, the control
included, at $6.4$--$9.0\times$ its selection-corrected floor. What produces the later readable form
at the queried position is attention-mediated gathering inside a mid-depth window: separating patch
depth from readout depth puts transport there at least $17\times$ above anywhere shallower under
non-saturating readouts, with no
tested MLP output contributing positively inside it. Under the saturating percentile rank the same
grid does not localise the window, which is a fact about that measure. An arm that needs the
variable for nothing concentrates sevenfold less, so the window is demand-specific.
That window has two measured edges, a survival failure
below and destruction above, and it falls at the same \emph{fractional} depth in a 64-layer hybrid
and a 62-layer dense model from another family. We localise where the variable is installed and read,
not the route from the passage, which transports nothing. But the readout
is not a calibrated
measure of use: three components move it to within $12\%$ of one another and differ $7.4\times$ in
what they do to the answer.
\end{abstract}

\section{Introduction}
\label{sec:intro}

Recent work identifies a set of directions in a language model's residual stream whose contents the
model can report on, a \emph{verbalizable workspace}, and shows that a latent quantity is more
present there when the task requires using it flexibly \citep{gurnee2026workspace}. What causes a
representation to enter that form is left open.

\begin{figure}[t]
\centering
\includegraphics[width=0.88\textwidth]{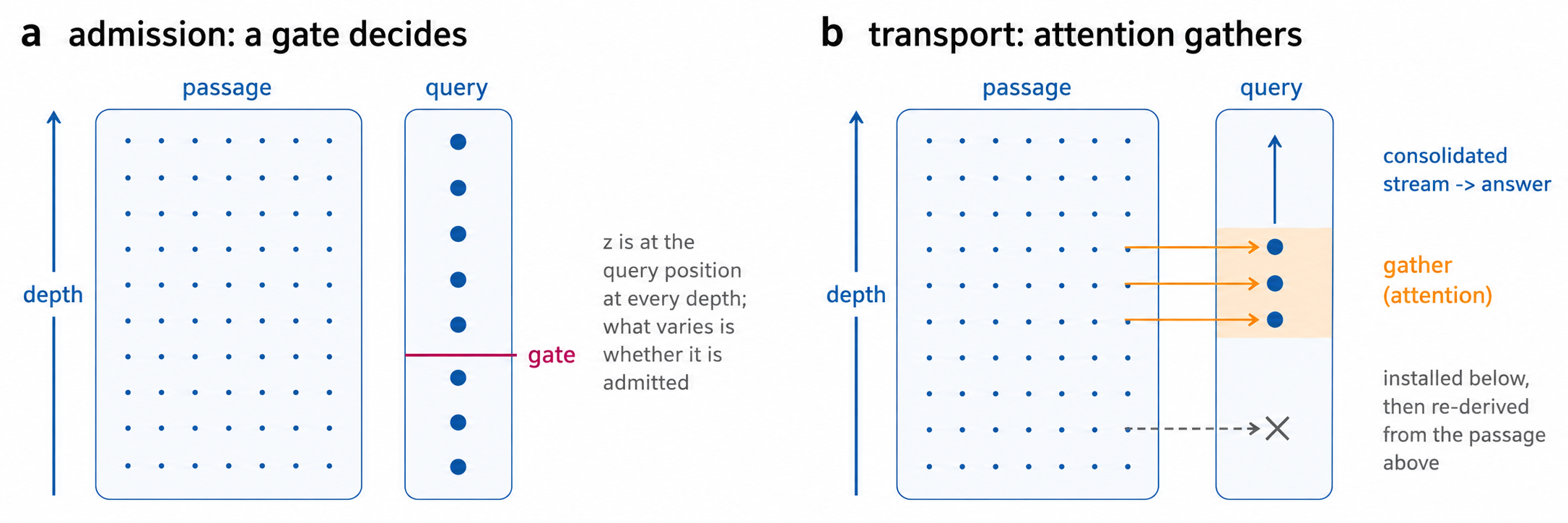}
\caption{Two accounts of what changes when a task demands flexible reuse of a latent variable $z$.
\textbf{(a)} Admission, as the workspace vocabulary invites one to read it: $z$ is at the query
position at every depth and a gate sets whether it enters. This is our formalisation of the
hypothesis, not a claim \citet{gurnee2026workspace} make. \textbf{(b)} What we measure: attention
gathers $z$ into the query position within a mid-depth window, a value installed below it does not
survive to the readout, and the answer depends far more on the consolidated stream than on the
gather. The passage is the plausible source but we do not localise it. (b) displaces (a) at the
query position; it does not rule out a gate acting on the gather itself. The re-derivation labelled in
(b) is our favoured reading of the lower edge, not a measurement (\S\ref{sec:window}).}
\label{fig:overview}
\end{figure}

The vocabulary of the finding suggests an answer. A workspace has contents; contents are
\emph{admitted}; admission implies a decision taken over something already present. Read that way
(Figure~\ref{fig:overview}a), the latent variable $z$ sits at the position where the model will be
asked about it, at every depth, and what demand changes is only whether some component lets it be
read. The reading is natural enough that concurrent engineered systems build it: an external
controller that writes into the late band when next-token entropy crosses a threshold
\citep{sathish2026recognition}, or an architectural gate trained to route information into a
workspace \citep{workspacerouting2025}.

An unmodified pretrained model does not have one there. Early at the query position, at least
seventeen times less of the variable is transportable than inside a mid-depth window, and across the
twelve layers immediately below that window no component transports it in every donor pairing. Attention
carries it in; no
tested MLP output inside the window contributes positively. And neither edge of the window is a
capability appearing: below it an installed value fails to survive to the readout, above it the
intervention stops substituting and starts destroying. So the picture is
Figure~\ref{fig:overview}b, and what moves with task demand is transport into the position where
the measurement is taken, not the unmasking of a variable that was sitting there all along.

\paragraph{Contributions.}
\textbf{(1)} Visibility dissociates from availability: demand raises it against a control matched
on prompt format \emph{and} accuracy, one shared linear map reads $z$ from every arm including the
one that needs it for nothing, and the effect survives holding operator use fixed on four
checkpoints from four families (\S\ref{sec:entry}). \textbf{(2)} The mechanism is transport by
attention. That attention rather than an MLP moves a value into the queried position is
\emph{convergent} \citep{geva2023dissecting,todd2024functionvectors}; the increment is the
\emph{dependence}, since our arms share an identical context and the amount transported changes
with what the task asks about a value that context never names (\S\ref{sec:gather}).
\textbf{(3)} Both edges of the window are measured rather than asserted, it lands at the same
fractional depth on a second architecture, and the behaviour runs through one derived lens
direction against four controls (\S\ref{sec:window}, \S\ref{sec:mechanism}). \textbf{(4)} We identify two nuisances that decide what a patching result is worth, neither
reported in the work we build on (\S\ref{sec:pairing}), and give a methodological warning with a
matched positive control (\S\ref{sec:illusion}).

We do not claim a write circuit. What is localised is a gather, and it is selectively necessary
without being sufficient.

\section{Setup}
\label{sec:setup}

\subsection{The lens, and what may be measured on it}
\label{sec:invariance}

The Jacobian lens summarises the map from an intermediate residual stream to the final one by one
averaged linear operator per layer, $J_\ell = \mathbb{E}[\partial h_{\text{final}}/\partial h_\ell]$,
and reads a layer out as $\mathrm{softmax}(W_U\,\mathrm{norm}(J_\ell h_\ell))$; $J_\ell = I$
recovers the logit lens. \emph{J-space} is the set of points expressible as a sparse nonnegative
combination of vocabulary-indexed lens vectors \citep{gurnee2026workspace}: a union of cones
rather than a subspace, which is why no subspace language appears below. We reserve the term for
that object; our dependent variable is the rank of a concept under the readout, which we call
\emph{lens visibility}. The two are related but not the same, and we checked rather than assumed how
closely: the sparse nonnegative coefficient that actually encodes membership correlates with our
readout at Spearman $0.24$ and is exactly zero in $90.9\%$ of activations, which is why it is not the
dependent variable and why we say \emph{visibility} throughout rather than membership
(Appendix~\ref{app:gauge}).

One class of quantity has to be excluded first. For any positive $a_\ell$ there is a network
computing the \emph{same function} whose residual stream is $a_\ell h_\ell$, and under it
$\tilde J_\ell = (a_{L-1}/a_\ell) J_\ell$. So anything computed on $J_\ell$ \emph{as a matrix} and
homogeneous of nonzero degree moves under a change the model cannot detect, and supports no claim
here; the readout itself does not move, since the two differ by a scalar the normaliser removes. The
rescaling must vary with depth to say anything at all, a global one leaving $J_\ell$ untouched. The
derivation, the fp32 check and the candidate checkpoint this disqualified are in
Appendix~\ref{app:gauge}.

Every dependent variable is therefore read after the model's own final normaliser. The primary one
is the percentile rank of the gold concept in the vocabulary $V$,
\begin{equation}
\Rz(h_\ell, z) \;=\; 1 - \frac{\bigl|\{\,v \in V : \lambda_v > \lambda_z\,\}\bigr|}{|V| - 1},
\qquad \lambda \;=\; W_U\,\mathrm{norm}(J_\ell h_\ell),
\label{eq:rz}
\end{equation}
a percentile rather than a raw rank so it compares across vocabulary sizes. It also
\textbf{saturates}, and not harmlessly: it separates rank $1$ from rank $25$ by $0.0001$, and
$92\%$ of flexible-arm cells at L40--L44 read above $0.999$, so a depth profile drawn from it alone
is compressed exactly where the effect is largest; Appendix~\ref{app:example} makes that concrete
on one instance. From the same rank computation we therefore also
record $\Lz = -\log_{10}(1 + |\{v : \lambda_v > \lambda_z\}|)$, which spends its resolution near the
top of the vocabulary, and $\Mz$, the log-probability margin against frequency-matched controls, and
report every transport result under all three. They place the depth peak about fifteen layers apart,
and the $\Rz$ and $\Lz$ depth profiles correlate at only $r = 0.581$, so a conclusion holding under
one and not another is a fact about the measure, and we say which.

\subsection{Checkpoints, and the benchmark}
\label{sec:bench}

The primary checkpoint is Qwen3.6-27B: 64 layers, $d = 5120$, vocabulary $248{,}320$, published
lens over all 63 source layers. It is a \emph{hybrid}: only every fourth layer carries standard
attention, with 24 heads, the other 48 a gated delta net. That governs what a ``head'' means
here and is load-bearing in \S\ref{sec:gather}. Instruction-tuned checkpoints throughout, since the
arms differ only in what their instruction asks for. Four more test how
far each claim travels: Qwen3.5-9B, phi-4 and gemma-4-31B-it on lenses \emph{we fitted}, and
Llama-3.1-8B-Instruct on a published one. Each cleared a tokenizer screen and, where a lens claim
rests on it, the homogeneity check above; two further candidates failed those screens
(Appendix~\ref{app:across}).

Each semantic instance of \textbf{JGateBench} yields five prompts over an \emph{identical} context,
so every contrast is within-instance and paired. \emph{Automatic} needs $z$ but never exposes it;
\emph{report} asks for $z$; \emph{flexible} passes $z$ through an operator defined in the prompt;
\emph{control} is a format-matched instruction that needs $z$ for nothing; \emph{supplied} applies
the same operator to a value \emph{given} in the prompt. The primary contrast is flexible against
control, which carries the identical \texttt{"Answer:"} format and no demand for $z$. The main
family reads language identity off FLORES-200 \citep{flores200}, whose $N$-way parallelism lets a
counterfactual vary the latent variable and nothing else; the second tracks objects under swaps,
where $z$ is a progressively updated state. Behaviour is scored throughout as a forced choice among
the four candidate answers at the final position, two in the yes/no control arm, never as an
open-vocabulary argmax.

Five design choices are invariants rather than preferences (Appendix~\ref{app:bench}). The
load-bearing one is \textbf{label symmetry across arms}: a flexible-arm operator table naturally
prints \texttt{"Spanish -> 7"}, putting the gold label in one prompt and not another, so the
contrast partly measures whether the label was printed. We share the table across every arm and
enforce the symmetry in code, because deliberately breaking it inflates the headline effect by
$26\%$: a large, clean, artifactual result in the predicted direction.

\section{Demand changes visibility, not availability}
\label{sec:entry}

\paragraph{Visibility rises with demand, at matched accuracy.}
Table~\ref{tab:entry} gives the paired contrasts over 200 semantic instances. Against the
format-matched control, lens visibility rises by
$\Delta\Rz = \ci{+0.0891}{+0.0799}{+0.0983}$, and by $\ci{+2.79}{+2.57}{+3.01}$ in the
non-saturating margin, while behavioural accuracy is matched to within noise of \emph{exactly} zero,
both arms at $0.940$, so the effect is not a difficulty difference wearing a representational
costume. The match is on raw accuracy: the control question is binary, so above their own floors the
arms read $0.920$ and $0.880$, a residual well inside the interval the match is quoted with. One
measurement choice matters later. We average $\Rz$ over the workspace band (L24--L59, the
depth-verified band of the prior survey) rather than
maximising it, and the profile that mean summarises peaks at L26 under $\Rz$ and at \textbf{L41}
under $\Lz$, where it reaches $\ci{+3.50}{+3.36}{+3.63}$ decades against $+1.22$ at L26,
\emph{inside} the window
\S\ref{sec:gather} localises causally, where the percentile scale had put it ten layers below.

\begin{table}[t]
\centering
\footnotesize
\setlength{\tabcolsep}{2.5pt}
\begin{tabular}{lccc}
\toprule
contrast & $\Delta\Rz$ & $\Delta\Mz$ & $\Delta$ accuracy \\
\midrule
\textbf{flexible $-$ control} & $\ci{+0.0891}{+0.0799}{+0.0983}$
  & $\ci{+2.79}{+2.57}{+3.01}$
  & $\ci{+0.000}{-0.045}{+0.045}$ \\
report $-$ control & $\ci{+0.0991}{+0.0891}{+0.1092}$ & $\ci{+3.52}{+3.31}{+3.73}$
  & $\ci{+0.040}{+0.005}{+0.075}$ \\
\textbf{flexible $-$ supplied} & $\ci{+0.0504}{+0.0445}{+0.0567}$
  & $\ci{+4.05}{+3.81}{+4.28}$ & $\ci{-0.060}{-0.095}{-0.030}$ \\
supplied $-$ control & $\ci{+0.0387}{+0.0310}{+0.0466}$ & $\ci{-1.26}{-1.37}{-1.15}$
  & $\ci{+0.060}{+0.030}{+0.095}$ \\
\bottomrule
\end{tabular}
\caption{Lens visibility by condition pair on Qwen3.6-27B, under the saturating percentile rank and
the non-saturating margin, both averaged over the workspace band. Paired bootstrap over semantic
instances ($10{,}000$ resamples,
$n = 200$), the instance being the resampling unit because all five arms share a passage. Arm
accuracies are \emph{control} $0.940$, \emph{report} $0.980$, \emph{flexible} $0.940$,
\emph{supplied} $1.000$; \emph{control} is a yes/no question where the others offer four candidates,
so measured above its own chance floor it reads $0.880$ against \emph{flexible}'s $0.920$. The first
row is matched on prompt format and on raw accuracy; the last two separate the operator from the
latent variable, and the operator's own contribution is the one quantity here that changes sign with
the measure.}
\label{tab:entry}
\end{table}

\paragraph{The variable is there either way.}
A multinomial probe on the residual stream at the query position, trained and tested on
\emph{disjoint semantic instances}, decodes $z$ far above chance in every arm. It is a \emph{single}
weight matrix fitted jointly across the four arms with no per-arm adjustment, and we quote the nested
form, choosing the layer on an inner split of the training instances only: $0.575$ in control to
$0.808$ in report over five split seeds, against a floor built by permuting labels under the
identical rule, which reaches $0.090$ where nominal chance is $0.05$, so $6.4$--$9.0\times$ the
floor. One matrix serving all four arms gives the strong form: the same linear map reads $z$ whatever
the task asks, including in the arm that needs it for nothing. The dissociation is graded rather than
absolute, since the within-arm diagonal itself rises with demand (Figure~\ref{fig:entry}).

\paragraph{Holding the operator fixed, on four checkpoints.}
Matching format and accuracy does not make the arms computationally equivalent: \emph{flexible}
must infer $z$ \emph{and} apply an operator to it, where \emph{control} does neither. The
\emph{supplied} arm separates the two by handing the operator a value in the prompt: a \emph{dummy} candidate rather than the true one, since naming the true one would pin its rank at
$\sim 1$ by copying. With operator use held approximately fixed, needing to infer the value adds
$\ci{+0.0504}{+0.0445}{+0.0567}$ on Qwen3.6-27B, and $\ci{+4.05}{+3.81}{+4.28}$ in the margin,
the largest effect in Table~\ref{tab:entry}. The same contrast is positive on every
checkpoint we ran it on: $\ci{+0.0429}{+0.0371}{+0.0493}$ on phi-4,
$\ci{+0.0346}{+0.0288}{+0.0408}$ on Llama-3.1-8B and $\ci{+0.0855}{+0.0743}{+0.0967}$ on
gemma-4-31B (Figure~\ref{fig:across}c), and positive on all four under the non-saturating readout
too, though it reorders them (Appendix~\ref{app:across}). Four architectures and four lenses, two of
them fitted here.

The \emph{operator's own} contribution is a warning rather than a result, and we report it as one.
It runs $+0.039$, $+0.178$, $+0.029$ and $\mathbf{-0.059}$ on those same four checkpoints, not stable in sign. It reverses on the primary checkpoint between the two readouts,
$+0.0387$ in percentile rank against $\ci{-1.26}{-1.37}{-1.15}$ in the margin, and it tracks per-arm
accuracy rather than the model (Appendix~\ref{app:across}). The decomposition needs accuracy-matched
arms, we have them on Qwen3.6-27B alone, and the split is scoped there.
Nor is it a variance decomposition even so: the interaction is significantly negative,
$\ci{-0.0487}{-0.0571}{-0.0405}$, so entering the table by the other path makes inference alone
account for $111\%$. The four conditional effects are the quantities that mean something.

\paragraph{Replication.}
The effect survives a change of task family (tracking gives flexible $-$ control
$\ci{+0.1792}{+0.1625}{+0.1961}$, twice the language effect) and a change of model \emph{and}
lens, at $\ci{+0.2931}{+0.2775}{+0.3084}$ for report $-$ control on Qwen3.5-9B. That second one runs
through \emph{report}, whose gold concept is also its next token, so part of its visibility is
imminent production rather than availability; the flagship contrast is immune, since its answer is an
operator symbol. Each needed its own
scope statement, and the two arrived from opposite directions (Appendix~\ref{app:tables}).

\section{The variable is transported, not unmasked}
\label{sec:gather}

Everything below substitutes a donor's activation and reads the concept through the lens, so the
quantity is \emph{donor-substitutable, lens-readable} content at the query position. A weak patching
effect is therefore not an absent representation: an early encoding that is nonlinear,
lens-misaligned or destroyed by substitution would read as weak. What we establish is the
narrower thing the admission account denies: that the later readable, causally substitutable form at
that position is produced predominantly by gathering rather than by unmasking a representation
already in place. Readout depth is a second axis and not a setting. Fixing it late grades
\emph{survival}, since a shallow patch is then measured after many layers in which the model can
overwrite it, while a fixed short distance grades installation; conflating the two produces a
spurious onset, and we produced one before separating them.

\begin{figure}[t]
\centering
\includegraphics[width=\textwidth]{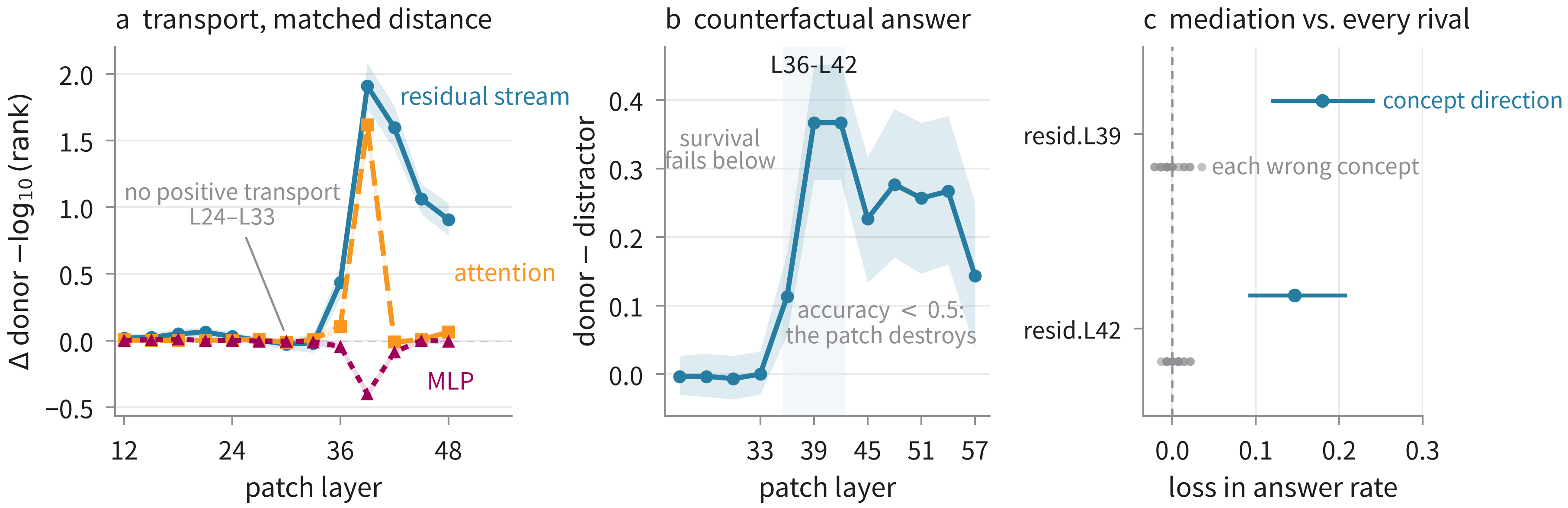}
\caption{The causal chain, measured three ways on the language family; all intervals are paired
bootstrap intervals over instances. \textbf{(a)} Where a donor's value can be \emph{installed},
read three to five layers above the patch so every depth is measured at matched distance, in the
non-saturating $\Lz$ at $n = 120$, averaged over four donor pairings; per-cell means, and how many pairings each clears,
are in Table~\ref{tab:transport}. \textbf{(b)} The behavioural counterpart
at $n = 150$: donor-symbol rate minus the matched per-distractor rate. The shaded band is where the
patch still leaves the task solvable; the effect is significant above it too, at accuracies of
$0.17$--$0.21$. \textbf{(c)} Specificity of the mediating direction at $n = 150$, against a null of
one point per wrong concept.}
\label{fig:spine}
\end{figure}

\paragraph{Transport is concentrated in the window, by at least seventeenfold.}
Read three to five layers above the patch, the depth profile is sharply peaked
(Figure~\ref{fig:spine}a). Over \emph{four} donor pairings the L39 stream carries
$24.8$--$37.8\times$ the largest cell anywhere below the window under $\Lz$, the weakest of the four
intervals still reaching $17.0$, and $51.6$--$62.6\times$ under $\Mz$; the shallow maximum is taken
inside each resample, so the ratio is conservative. Under $\Rz$ it is $1.1$--$2.7\times$ with every
interval covering $1$: the percentile scale compresses the window's own effect, which is the
sharpest case of the saturation \S\ref{sec:setup} documents and the reason no depth claim here rests
on it. Below the window only two of twenty-four cells clear zero in all four pairings, both in the
residual stream and neither in attention or MLP, and across L24--L33 not one of twelve transports
positively under either rank readout, with intervals excluding $\Rz$ effects above $0.0044$.

\paragraph{The window is demand-specific, not a property of the geometry.}
The passage sits in the same place in every arm, so a window arising only from where the readout is
taken should appear in an arm that needs the latent for nothing. It appears, and it is much flatter:
on the same donor pairing the \emph{control} arm concentrates at $4.2\times$ $[2.9, 6.6]$ against
\emph{flexible}'s $30.3\times$ $[18.6, 53.4]$, intervals not overlapping. The difference sits in the
attention branch at the gathering layer, which transports $16.5\times$ as much under demand ($+1.610$
against $+0.098$ in $\Lz$), where at the window's lower edge the two arms are within a fifth of each
other. Some of the variable travels whatever the task asks; what demand changes is the gather at the
peak.

\paragraph{Attention carries most of it; no MLP in the window carries any.}
At L39 the attention output and the residual stream are indistinguishable under $\Rz$
($+0.0135$ against $+0.0137$), but that near-equality belongs to the saturating metric, and $\Lz$
separates them at $+1.618$ against $+1.908$, so attention carries $85\%$ of the stream's effect
rather than all of it. Independent patch effects are not additive in a nonlinear network, so neither
figure proves it carries the whole; separating them needs path patching
\citep{goldowskydill2023pathpatching}, which we have not run. Nor
is this a property of full-attention layers as a class: L15 and L27 are also full-attention and
neither transports significantly. Meanwhile \texttt{mlp.L39} \emph{opposes} the transport at
$-0.0301$ under $\Rz$ and $-0.400$ under $\Lz$, the grid's largest negative cell under both readouts
and negative in all four pairings (Table~\ref{tab:transport}). The second family is drawn in
Figure~\ref{fig:gather} and this one in Figure~\ref{fig:spine}a. Only
attention can route content \emph{between} positions, so the MLP cells test the narrower question of
whether a local MLP output independently carries donor-aligned content, and inside the window they do
not. That much is convergent, and \citet{gurnee2026workspace} already report heads relaying J-space
content between positions; our increment is the \emph{dependence}, since the arms share an identical
context and the amount transported changes with what the task asks.

\paragraph{The second family replicates the branch and spreads the gather.}
On tracking at $n = 140$, under four donor pairings, \texttt{resid.L39} transports
$+0.0346$ to $+0.0505$, surviving Benjamini--Hochberg over 42 components in all four; so do the
attention blocks at L39 and L48 and the prespecified head \texttt{attn.L39.H15} at a fifth of the
stream; \texttt{mlp.L39} is negative in all four. What does \emph{not} carry over is the
concentration: on language \texttt{attn.L48} is a twenty-fifth of \texttt{attn.L39} under $\Lz$ and
one head matches the whole block, where on tracking the two blocks are comparable and no head
dominates. L48 is a \emph{linear}-attention layer, so no head decomposition is possible there
(Appendix~\ref{app:transport}).
\citet{gupta2026tabpfn} report head dominance shifting with task complexity in an unrelated model
class. Outside the window the largest positive MLP point estimate is \texttt{mlp.L15}, which no
pairing resolves from zero, so we scope the claim to the window rather than asserting it of MLPs in
general. The defensible claim is: \emph{a task latent
variable is gathered into the query position by attention within a mid-depth window, inside which no
tested MLP output contributes positively; how the gather is shared across that window's attention
layers is task-dependent.}

\section{The window has two measured edges}
\label{sec:window}

Target and donor are both flexible records with their own randomised operator tables, so a patch
carries a \emph{specific} predicted wrong answer: the target's table applied to the \emph{donor's}
language. That symbol appears nowhere in the donor's prompt, and producing it requires the target's
table to be applied \emph{after} the transport: compositional rather than copying. The control is
an identity. With $C$ the candidate set, $s_d$ the donor's predicted symbol and
$D = C \setminus \{s_{\text{gold}}, s_d\}$ the distractors, we report
\begin{equation}
\hat\Delta \;=\; \mathbb{E}\bigl[\mathbf{1}\{a = s_d\}\bigr]
\;-\; \mathbb{E}\Bigl[\tfrac{1}{|D|}\textstyle\sum_{s \in D}\mathbf{1}\{a = s\}\Bigr].
\label{eq:distractor}
\end{equation}
Under the null that a patch destroys the computation and leaves the answer uniform on $C$, both
terms equal $1/|C|$ and $\mathbb{E}[\hat\Delta] = 0$. Comparing against the clean run instead has
expectation $0.25$ here, which is how one of our own runs produced six confident false positives.

\paragraph{The window.}
At $n = 150$ the counterfactual is null through L33, becomes detectable at \textbf{L36}
($\ci{+0.113}{+0.057}{+0.177}$, task accuracy $0.807$) and plateaus over L39--L45 at $+0.227$ to
$+0.400$ across three pairings (Figure~\ref{fig:spine}b). Neither edge is a step, and we name no peak: across pairings L39,
L42 and L45 trade places. Significance runs
on to L57, but accuracy has fallen from $0.920$ to $0.19$ by L48, so the interpretable cells are
L36--L42; deeper, a patch replaces the late computation wholesale and the donor's symbol
arrives without anything having been transported. The band is narrower than that reads, and we state
it plainly: L36 is the only cell with comfortable accuracy and it is the one cell the adversarial
destruction bound of \S\ref{sec:pairing} does not clear, while L39 and L42, which do clear it, sit at
$0.553$ and $0.533$, just above the floor at which we ourselves stop calling a patch a
substitution. Patched-position count is an axis too: the final
query token alone leaves the flip rate within noise,
$\ci{+0.0250}{-0.0250}{+0.0813}$ against $\ci{+0.3375}{+0.2188}{+0.4562}$ over twelve, while the
graded margin there clears zero, $\ci{+0.550}{+0.271}{+0.831}$, so that position
receives transport without the choice flipping, and the behavioural claim concerns a query
\emph{span} read at its final token (Figure~\ref{fig:window}, Appendix~\ref{app:behaviour}).

\paragraph{The lower edge is survival, and repair is the likeliest reason.}
Because only the last tokens are patched, the passage stays the target's, and layers above a patch
can attend back to it and re-derive the target's own value. Patching \texttt{resid} at L33--L45 and
reading the \emph{donor's} value at a fixed later band measures survival directly: a value installed
at L33 is entirely gone, $\ci{+0.0048}{-0.0010}{+0.0113}$; at L36 about half survives,
$\ci{+0.0453}{+0.0326}{+0.0596}$ against the $+0.0888$ asymptote; from L39 it saturates. So the
lower edge is a survival boundary: the content is transportable earlier, it just does not last.
\emph{Why} is an inference and we mark it as one. Re-derivation from the still-present passage is the
reading we favour, and the fact that survival saturates from L39 rather than growing with the number
of layers left is a point in its favour; but overwriting, attenuation through the normalisers, and
plain incompatibility with the surrounding activations all predict the same disappearance. Deciding
between them needs the target's passage masked after an early patch, which we did not run
(Appendix~\ref{app:behaviour}).

\paragraph{No tested set of component writes reconstructs the stream's effect.}
Patching \emph{every} attention and MLP output across L36--L42 (fourteen components,
Table~\ref{tab:joint}) reaches $\ci{+0.0250}{-0.0167}{+0.0833}$ against \texttt{resid.L42}'s $\ci{+0.4250}{+0.2833}{+0.5667}$ in the
same run. That is weaker than showing the writes are irrelevant, since simultaneously substituted
outputs may be mutually inconsistent in a nonlinear network, the interaction term
\citet{khemais2026collapse} derives in closed form. The answer depends on the accumulated stream as
we can measure it; the gather is necessary, not sufficient.

\subsection{The same window on a second architecture}
\label{sec:across}

Absolute layer indices are not comparable across checkpoints, so the cross-architecture unit is
fractional depth. On gemma-4-31B-it (62 layers, dense, another family, and a measurement that
never touches a lens) the sweep is null or unstable at or below fractional depth $0.55$ and becomes
interpretable at exactly one cell, L37 at $0.60$: $+0.170$, $+0.107$ and $+0.140$ under three donor
pairings, at task accuracy $0.74$ to $0.77$ (Figure~\ref{fig:across}b). Qwen3.6-27B's interpretable
cells sit at $0.56$--$0.66$. Above that, gemma-4 does what Qwen3.6-27B does above L45 only more
sharply: the effect jumps to $+0.55$--$+0.65$ while accuracy collapses to $0.09$--$0.15$, the
largest wholesale-copying signal we have measured and the best argument in the paper for reporting
accuracy beside every counterfactual rate. We did not run the lens-side sweep there, so the
attention/MLP asymmetry is untested on this checkpoint; Appendix~\ref{app:across} splits L37 by
component kind under one pairing.

A third checkpoint shows what the design needs. Llama-3.1-8B-Instruct solves the flexible arm at
$0.610$, and that is not enough headroom: every cell whose effect clears zero does so at a patched
accuracy of $0.34$ or below, and the only cells above $0.39$ are null. Nothing on it is interpretable
under the criterion we apply to the other two. That is a scope condition on the design rather than
evidence against the window: the sweep needs a checkpoint starting near ceiling, which is why
gemma-4-31B-it at $0.970$ resolves it and this one does not.

\section{What a single donor pairing is worth}
\label{sec:pairing}

Donors are drawn from the whole instance pool, so a different seed reassigns every pair. That seed
is a free design parameter, and on the tracking sweep it decides a great deal:
Benjamini--Hochberg at $q = 0.05$ returns $21$, $19$, $15$ and $23$ significant components of 42
under four pairings, and only $14$ survive all four. A single pairing therefore reports $7$--$64\%$
more significant components than are stable, depending on the draw, and the excess is almost entirely \emph{individual heads}: ten of the twelve unstable cells are heads of L39 at point estimates of
$\pm 0.001$--$0.005$, while stream-level and whole-block cells are stable. We know of no
activation-patching work that reports this sensitivity. It bounds what a head-level claim from one
pairing is worth, ours included, which survives because it exceeds every unstable cell in every
pairing. Four of our sweeps are reseeded and reported as ranges: tracking, the
counterfactual depth sweep, the mediation and the language matched-distance grid of
\S\ref{sec:gather}. The L39 head decomposition is single-pairing, and we say so wherever its
numbers appear.

The second nuisance is in Equation~\ref{eq:distractor}, which is unbiased under \emph{uniform} destruction; real destruction is not uniform. With the realised candidate set recorded per
trial the residual is exact, and its sign follows the pairing: $+0.0043$ under one and $-0.0154$
under another. So it is a $\pm 0.015$ nuisance of the draw rather than a bias of the instrument, and
the correct statement asserts no direction. Under an adversarial bound in which \emph{every} pair is
as bad as the worst one observed, L39--L45 survive on both pairings and L36 does not
(Appendix~\ref{app:behaviour}). What the control cannot do is separate transport from wholesale
copying; that guard is task accuracy, not the estimator.

\section{What the behaviour runs through}
\label{sec:mechanism}

\paragraph{Selective necessity.}
Ablating a component to a \emph{leave-one-out} per-condition mean (so no record is ablated partly
towards itself) and comparing the accuracy cost in the arms that need $z$ against the
format-matched control gives a selectivity measure. The strongest dissociation is
\texttt{resid.L39}: report loses $\ci{-0.600}{-0.740}{-0.460}$ ($0.980 \to 0.380$) and flexible
$\ci{-0.120}{-0.220}{-0.040}$, while the control arm \emph{improves} by
$\ci{+0.080}{+0.020}{+0.160}$. It is the only component clearing zero here;
\texttt{attn.L39.H15} points the same way in the flexible arm alone and does not,
$\ci{-0.060}{-0.140}{+0.000}$. The control-arm nulls are not a measurement floor: the harsher
\emph{zero} ablation of \texttt{mlp.L39} costs \emph{control} $\ci{-0.220}{-0.340}{-0.120}$ and
leaves the other two arms at exactly $0.000$ (Table~\ref{tab:ablate}).

\paragraph{One derived direction carries part of it.}
A residual patch carries everything the donor computed, so the effect might run through the
concept's lens component or through something else travelling with it; the test removes the
candidate mediator and keeps the rest. The direction is derived rather than fitted. Since the
readout is $W_U\,\mathrm{norm}(J_\ell h_\ell)$, the vector $J_\ell^\top W_U[z]$ is the \emph{static}
pre-normaliser lens vector for token $z$, not the exact logit gradient, whose projection under RMSNorm
removes the component along the activation and so removes identically none of it
(Appendix~\ref{app:behaviour}). Projecting the static vector out of a $5120$-dimensional residual
costs $+0.113$ to $+0.193$ of the counterfactual answer rate at L39 and $+0.080$ to $+0.147$ at L42
across three pairings, $19$--$46\%$ of the rate on the same trials, with task accuracy recovering
correspondingly.

Four controls run on those same pairs, tabulated with the L42 cell in
Appendix~\ref{app:behaviour}: a random direction rescaled to strip \emph{exactly} as much activation
norm costs nothing ($-0.013$ to $+0.007$), and the mean over rival concepts is null ($-0.017$ to
$+0.015$), while the gold direction costs $+0.113$ to $+0.193$ and survives orthogonalising against
the other 19 ($+0.127$ to $+0.213$); the worst of twenty rivals reaches $+0.014$ to $+0.042$.
Two nulls and two positives make the effect hard to attribute to the projection rather than to the
concept (Figure~\ref{fig:spine}c). The wrong-concept null is a distribution rather than a draw: each pair runs against the 19 languages that are not its donor, pooled into one estimate per concept identity, and its worst cell is a maximum over twenty, hence the range. Because the 20 concept
directions have mean pairwise cosine
$0.50$, projecting out the gold one also strips most of every other; orthogonalising against the
remaining 19 leaves the effect intact, as does the surgical variant that removes only what the
substitution introduces. One limit remains: at least half survives removal, so something else in the
stream contributes and we do not identify it, which is why we say the behaviour \emph{runs through}
the direction rather than that it is mediated by $z$ (Figure~\ref{fig:mechanism}).

\section{A readout shift is not a measure of use}
\label{sec:illusion}

The clearest single fact here is negative, and about method rather than this model. At
L39, patching the attention output from a mismatched donor moves the lens readout toward the donor's
language by $\ci{+0.1333}{+0.0871}{+0.1847}$, $4.9\times$ the shift in the target's own language on
the same patched trials, and the same component is the only one whose effect reverses with the
direction of substitution (Appendix~\ref{app:tables}). It nonetheless newly selects the predicted
counterfactual answer under forced choice in
\textbf{$1$ of $80$ pairs}, with its \texttt{H15} slice in none, where the
residual stream at the same layer (same donors, same positions, same pairs) selects it in $29$.
A thresholded flip rate cannot tell an inert patch
from one that moves the answer without overtaking it, so we measured behaviour continuously: the
attention patch is \emph{not} inert, raising the donor symbol's margin over the distractors
significantly, about a seventh as far as the stream does.

On the same trials the three components move the donor's readout in $\Lz$ by $+1.83$, $+1.68$ and
$+1.63$, within $12\%$ of one another, while their effect on the donor-minus-distractor margin
differs $7.4\times$, at $+2.56$, $+0.44$ and $+0.35$ (Figure~\ref{fig:calibration}). Trial-level
correlations between the two run $+0.27$, $+0.18$ and $+0.08$: weak, but for two of the three
genuinely nonzero. So the readout is not uninformative about behaviour; it is \emph{uncalibrated}:
its magnitude cannot be read as causal influence when three components at the same readout level
do three very different things. That component patching can activate pathways disconnected from the
output is known \citep{makelov2023illusion,zhang2024patching}. What we add is a controlled instance
on the Jacobian lens, where that inference is routinely made, with a
matched \emph{positive} control: \S\ref{sec:mechanism} shows the concept-aligned component is not
epiphenomenal in general. The stakes are not hypothetical: \citet{prosvirnin2026jadr} score models
for safety from top-$k$ J-space tokens per layer; our result motivates a causal step
before any readout difference is read mechanistically.

\section{Related work}
\label{sec:related}

\citet{gurnee2026workspace} establish the automatic/flexible distinction, the lens and J-space, and
report attention heads relaying J-space content between positions, so the mechanism is not open in
general and we do not claim it is. They leave open what causes a representation to \emph{enter} the
workspace. Four differences make our transport result an increment: their relay heads are selected
from weights and ablated at every token position, so relay is not separated from same-position
writing; their relay analysis and demand contrast are never crossed; readout depth is never an
independent axis, so installation is not separated from survival; and their causal results are on
closed models.

Concurrent systems build the gate we went looking for
\citep{wu2026jcot,sathish2026recognition,workspacerouting2025}: construction against discovery, and
our answer is that their gate is not where the architecture puts it, because there is no decision to
take over a variable not yet at the readout position. Our gather belongs to an established literature
on attention moving content into the prediction position
\citep{geva2023dissecting,wang2023ioi,todd2024functionvectors}; it differs in that the variable's
\emph{value} is never named, so there is no source token to copy, and the amount transported moves
with demand, not the prompt. Our depth-swept patching follows \citet{meng2022rome} and
\citet{vig2020causal}. Our latent is language identity, studied by \citet{wendler2024llamas} and
\citet{dumas2025tongue}; we hold the passage fixed and ask what makes it readable at the queried
position. Closest in shape to our
negative result, \citet{bersia2026oracles} find concept-specific blind spots where the target
``remains decodable inside the oracle''. Adjacent work covers attention-gated routing, causal head
gating, intervention mechanics and course-correction under steering
\citep{alignmentroutes2026,headgating2025,promptactivation2026,esr2026}.

\section{Limitations}
\label{sec:limitations}

What we rule out is an admission step that unmasks a variable already sitting at the query position.
A gate on the attention \emph{route} stays consistent with everything here: passage-attention mass at
the gathering layer is $1.90\times$ as demand-sensitive as at the 72 heads of three non-gathering
layers, median against median, though maximum against maximum is only $1.16\times$, and it is
measurable only on full-attention layers (Appendix~\ref{app:transport}). So the honest statement is
narrower than our title: we refute admission \emph{at the query position}, and what sets the amount
transported is plausibly a gate on the route we do not identify.

On scope, the entry effect is measured on five checkpoints and two task families; the mechanism on
far less: the window on two architectures, and the attention route, the head, the mediation and the
ablations on Qwen3.6-27B and the language family alone. We localise the gather but not the computation
feeding it: no passage span we patched transports the variable, leaving its pre-gather form
unlocalised rather than distributed. The gather is necessary without being sufficient, with
mediation partial. And \emph{supplied} sits at accuracy $1.000$ against \emph{flexible}'s
$0.940$, a ceiling in the $2\times2$ that no analysis removes.

\section{Conclusion}

An admission account presupposes a step that does not happen where the readout is taken: the variable
is decodable from every arm, and the measurement sits where it has to be \emph{moved} to. Attention
does the moving, necessarily but not sufficiently, at one fractional depth in two architectures and
sevenfold sharper under demand. And even there, the size of a readout shift says nothing about what
a component does to the answer.

\subsection*{Reproducibility statement}

Every number here is written by a command-line entry point into a JSON or JSONL artifact on disk;
none is transcribed. The benchmark generator, the five design invariants and their guard tests are in
Appendix~\ref{app:bench}; the checkpoint revisions, lens files, decoding settings and what is seeded
are in Appendix~\ref{app:artifacts}; the cross-architecture screens, including the two candidates
that failed them, are in Appendix~\ref{app:across}. Figures are built from artifacts by code rather
than transcribed, and an audit command matches every numeric literal in this manuscript against the
artifact tree. Where a quantity depends on the donor pairing we report its range over pairings and
name the sweeps for which we have only one (\S\ref{sec:pairing}).

\subsection*{AI use statement}

We used generative AI assistants to help with coding the project and for feedback on our experiment
design. We did not use them to interpret results, to develop or prove any mathematical claim, or to
generate any benchmark item. We have reviewed all AI-assisted work and take responsibility for the
final content of this paper.

\bibliography{main}
\bibliographystyle{iclr2027_conference}

\appendix

\section{The benchmark, and the invariants it enforces}
\label{app:bench}

\paragraph{One instance, all five arms.}
Instance \texttt{lang\_000169}, verbatim apart from the passage. Every arm receives the same prompt
up to and including the blank line, then one instruction:

\begin{quote}
\footnotesize
\begin{verbatim}
Reference table:
Finnish -> W
Japanese -> K
Swedish -> V
Turkish -> Z

<FLORES-200 passage, three sentences>

[instruction]
\end{verbatim}
\end{quote}

\noindent The five instructions, each followed by \texttt{\textbackslash nAnswer:}, with the gold
answer it is scored against and the size of its candidate set:

\begin{center}
\scriptsize
\setlength{\tabcolsep}{3pt}
\begin{tabular}{llc}
\toprule
arm & instruction & gold ($|C|$) \\
\midrule
\emph{automatic} & \texttt{Continue the passage. Write the next sentence.} & next sentence (4) \\
\emph{report} & \texttt{What language is the passage written in?} & \texttt{Japanese} (4) \\
\emph{flexible} & \texttt{What symbol corresponds to the language of the passage?} & \texttt{K} (4) \\
\emph{control} & \texttt{Does the passage contain a question mark?} & \texttt{No} (2) \\
\emph{supplied} & \texttt{Treat the language as Swedish. Which symbol is it?} & \texttt{V} (4) \\
\bottomrule
\end{tabular}
\end{center}

\noindent Four things in this one instance are the design: the reference table names \emph{every}
candidate symbol in every arm, which is label symmetry and is also why absolute $\Rz$ levels are not
unprompted representation strength; \emph{control} shares the \texttt{"Answer:"} format while asking
something the passage answers without its language being identified; \emph{supplied} names
\emph{Swedish}, a dummy and not the passage's actual \texttt{Japanese}, so its gold is \texttt{V}
rather than \texttt{K} and no copying route can produce it; and the passage is quoted, never
generated. The passage is elided here; it is recoverable from the released records by instance id.

\paragraph{Family sizes.}
The language family generates $1954$ records over $400$ semantic instances, of which $354$ are
complete in all five conditions; $46$ instances are dropped from the \emph{automatic} arm, the
limiting one, because two candidate languages share a continuation first token. The four-arm subset
used by every measurement other than the $2\times2$ is drawn from the same $354$ instances, $200$ of
them at the cap the entry measurements use, taken by sorted identifier so that two runs at one cap
see the same instances. Twenty distinct
languages survive filtering. Prompt lengths are $107$ / $140$ / $235$ tokens (min / median / max),
all clear of the lens's unfitted position floor. FLORES-200 supplies $1012$ sentences per language.

\paragraph{The five invariants, and the failure each prevents.}
\begin{enumerate}
  \item \textbf{Label symmetry across arms.} A flexible-arm operator table naturally prints the gold
    label, putting it in one prompt and not the other, so the contrast partly measures whether the
    label was printed and returns a large clean artifactual effect in the predicted direction. We
    share the table across every arm and check programmatically that every arm names the label or
    none does. Deliberately breaking it gives $\Delta\Rz = \ci{+0.1126}{+0.1015}{+0.1240}$ against
    the matched design's $+0.0891$: an inflation of $+0.0235$, about $26\%$ of the matched effect,
    with $79\%$ of the unmatched effect surviving the guard. The unmatched design also breaks
    accuracy matching ($1.000$ against
    $0.875$), so it is worse on two axes at once.
  \item \textbf{The label never appears in the passage.} A concept present in the context reads at
    rank ${\sim}1$ at its own position.
  \item \textbf{Forced choice, never open-vocabulary argmax.} On the primary checkpoint the top
    token is often \texttt{'\textbackslash n\textbackslash n'} with the gold answer at rank 1:
    $0.144$ against $0.955$ on the same trials.
  \item \textbf{Score the continuation form.} \texttt{" Spanish"} and \texttt{"Spanish"} are
    different token ids, worth $0/36$ against $10/36$ on a real checkpoint.
  \item \textbf{Single-token-ness is per-tokenizer and must be filtered.} On this tokenizer
    \emph{no digit} is single-token in continuation form (\texttt{" 7"} is two tokens), so numeric
    latents are unscoreable against a token-indexed lens here, which rules out any family whose
    answer is a number. All 20 language names pass; 9 of 18 operator symbols do, and all 9 survivors
    are letters.
\end{enumerate}

\paragraph{Statistics.}
All intervals are $95\%$ bootstrap intervals over $10{,}000$ resamples, clustered on the semantic
instance because the arms share a passage. Multiple comparisons are controlled within each sweep, over all of that sweep's
components, by Benjamini--Hochberg at $q = 0.05$: 39 cells for the language matched-distance sweep,
42 for tracking, 36 for the L39 head decomposition, 12 for each depth sweep. \textbf{Correction runs
independently inside each donor pairing}, because four pairings are four experiments rather than four
draws to average (\S\ref{sec:pairing}). The entry contrasts of Table~\ref{tab:entry}, the mediation,
the necessity table and the readout-against-use comparison are pre-specified single comparisons,
reported uncorrected. We give an absolute gap beside every ratio, and refuse ratios whose denominator
approaches zero.

\paragraph{Selection provenance, and a confirmation split.}
L39 and \texttt{H15} were chosen by looking at data, and it matters which data. The layer came from a
coarse screen of 26 components over \textbf{20} semantic instances, the head from a decomposition
over \textbf{25}. Both were frozen at that point and never revisited, and every downstream experiment
ran afterwards on strictly larger instance pools from the same family. The union of the two screens is
25 instances, so most of every downstream result rests on instances the selection never saw, which gives a confirmation split for free:

\begin{center}
\small
\begin{tabular}{lcc}
\toprule
 & selection pool ($n = 25$) & held out ($n = 55$) \\
\midrule
\texttt{resid.L39} counterfactual & $\ci{+0.400}{+0.200}{+0.600}$ & $\ci{+0.364}{+0.236}{+0.491}$ \\
mediation, \texttt{resid.L39} & $\ci{+0.120}{+0.000}{+0.280}$ & $\ci{+0.218}{+0.109}{+0.327}$ \\
mediation, \texttt{resid.L42} & $\ci{+0.080}{+0.000}{+0.200}$ & $\ci{+0.127}{+0.055}{+0.218}$ \\
\bottomrule
\end{tabular}
\end{center}

The counterfactual holds on unseen instances, and the mediation is \emph{stronger} there than on the
selection pool, where it does not clear zero at $n = 25$, so that result is carried entirely by
instances the layer choice never touched. Beyond the table, the \texttt{attn.L39} bidirectional
signature is slightly stronger held out ($+0.0572$ against $+0.0440$), necessity and the early-layer
cells hold at comparable or larger magnitudes, and \textbf{one secondary cell loses significance held
out} while the pre-L36 representational cells are underpowered at $n = 25$ rather than refuted.
\textbf{The head screen cannot be split, because it \emph{is} the selection pool.}
Selecting instances by sorted identifier correlates with nothing that matters: FLORES row index
($\rho = -0.078$), passage length ($\rho = -0.103$), gold language ($\chi^2$, $p = 0.87$), candidate
set or operator symbol. What the split cannot do is validate the choice of L39 itself: a different
screen might have picked a different layer, and \S\ref{sec:gather} shows the second task family does
exactly that.

\section{The gauge argument, and the sparse decomposition}
\label{app:gauge}

\paragraph{The reparameterisation, in full.}
Write $o_\ell$ for the output of block $\ell$ of a pre-norm residual transformer, so
$o_\ell = o_{\ell-1} + F_\ell(\mathrm{norm}(o_{\ell-1}))$. For any positive $a_\ell$ there is a
network computing the \emph{same function} whose residual stream is $\tilde o_\ell = a_\ell o_\ell$:
scale the embedding by $a_{-1}$, scale block $\ell$'s output projections by $a_{\ell-1}$, and
multiply the block's \emph{whole} output, skip included, by $a_\ell / a_{\ell-1}$. The two
factors compose, so $F_\ell$ is scaled by $a_\ell$ net, which is what makes the recurrence work:
\begin{equation*}
\tilde o_\ell
= \frac{a_\ell}{a_{\ell-1}}\Bigl[\tilde o_{\ell-1}
  + a_{\ell-1} F_\ell\bigl(\mathrm{norm}(\tilde o_{\ell-1})\bigr)\Bigr]
= \frac{a_\ell}{a_{\ell-1}}\Bigl[a_{\ell-1} o_{\ell-1}
  + a_{\ell-1} F_\ell\bigl(\mathrm{norm}(o_{\ell-1})\bigr)\Bigr]
= a_\ell o_\ell ,
\end{equation*}
using $\mathrm{norm}(a_{\ell-1} o_{\ell-1}) = \mathrm{norm}(o_{\ell-1})$. Every block still reads
through a scale-free normaliser, so it computes the same function of the same argument, and the
logits $W_U\,\mathrm{norm}(\tilde o_{L-1})$ are unchanged. Under this reparameterisation
\begin{equation}
\tilde J_\ell = \frac{a_{L-1}}{a_\ell}\, J_\ell .
\label{eq:gauge}
\end{equation}

Two scope notes the body compresses. The gauge \textbf{must vary with depth}: a global rescale scales
$o_\ell$ and $o_{L-1}$ alike, leaves $J_\ell$ untouched, and establishes nothing. And a depth-varying
$a_\ell$ needs a scalar on the residual path to absorb $a_{\ell+1}/a_\ell$, which a plain pre-norm
skip does not have, so the reparameterised network computes the same function, which is all the
argument needs, but it is not the same architecture, and we do not claim the gauge is realisable
in-architecture.

We verified the identity numerically rather than asserting it. On Qwen3.5-9B in fp32, with
$a_\ell = 1$ below layer 20 and $2$ at or above it, the largest change in any model logit is
$1.4 \times 10^{-5}$ and the largest change in $\Rz$ over 18 band layers is $4.0 \times 10^{-6}$,
both at fp32 rounding, while the mean diagonal of $J_\ell$ (by Hutchinson's estimator on
central-difference Jacobian-vector products) moves by exactly $2.0000$ at layer 19 and $1.0000$ at
layer 20, as Equation~\ref{eq:gauge} predicts. The diagnostic moves; the model and the dependent
variable do not. \textbf{The check has to run in fp32}: in bf16 the same construction drifts by
${\sim}3\%$ from accumulation alone, which blurs an exact identity into a small
residual and establishes nothing.

Two further notes. The lens readout is invariant \emph{without refitting} because $J_\ell h_\ell$ is
homogeneous of degree one; a learned affine translator is not, since its bias does not scale with
$h_\ell$, so one fitted in a given parameterisation must be refitted in another. And the closest
treatment we know of, \citet{belrose2023tunedlens} on drifting covariance and high-variance ``rogue''
dimensions, is empirical rather than an invariance argument. The screen is also load-bearing for checkpoint selection: it is what
excluded one candidate whose gated MLP is homogeneous of degree $1.76$--$1.93$
(Appendix~\ref{app:across}).

\paragraph{The sparse decomposition.}
Running full-dictionary gradient pursuit over all $248{,}320$ atoms at L39 without prefiltering the
vocabulary (support $k \leq 25$, a 24-second CPU computation, released with the artifacts) gives a
normalised gold coefficient $S_z$ correlating with $\Rz$ at Spearman $0.24$: a concept can enter the
support at rank $13{,}332$, and can rank third and stay out of it. This is an approximation, not a
certified global optimum (the problem is combinatorial). $S_z$ cannot serve as a dependent variable here: it is exactly zero in
$90.9\%$ of activations, and in the two arms whose $\Rz$ varies most it is nonzero in $4$ of $200$
(automatic) and $1$ of $200$ (control), so it supports no depth profile, no regression on layer and no
correlation with behaviour. It also moves by $9.1\times$ with the free parameter $k$. Support
\emph{membership} is better behaved, and the demand contrast is positive under it as under everything
else. We therefore measure visibility, say so, and report the decomposition as a check.

\section{The behavioural window, in detail}
\label{app:behaviour}

\paragraph{The window's cells, three donor pairings.}
The significance decision is the same in all twelve cells under all three pairings. Point estimates
move, so we give the range rather than one draw:

\begin{center}
\footnotesize
\setlength{\tabcolsep}{3.5pt}
\begin{tabular}{lccccc}
\toprule
patch layer & pairing 0 & pairing 1 & pairing 2 & worst bound & accuracy \\
\midrule
L24--L33 & $-0.007$ & $-0.017$ & $-0.020$ & all null & $0.91$--$0.93$ \\
\textbf{L36} & $+0.113$ & $+0.090$ & $+0.067$ & $+0.010$ & $0.807$ \\
L39 & $+0.367$ & $+0.300$ & $+0.400$ & $+0.220$ & $0.553$ \\
L42 & $+0.367$ & $+0.320$ & $+0.377$ & $+0.237$ & $0.533$ \\
L45 & $+0.227$ & $+0.330$ & $+0.337$ & $+0.133$ & $0.453$ \\
L48--L57 & $+0.143$ to $+0.277$ & $+0.230$ to $+0.293$ & $+0.150$ to $+0.320$ & $+0.037$ & $0.17$--$0.21$ \\
\bottomrule
\end{tabular}
\end{center}

\noindent ``Worst bound'' is the lower end of the least favourable of the three bootstrap intervals,
so a positive entry means the cell clears zero under every pairing. Two of the three pairings put L45
level with L42 and one puts it well below, so \textbf{no pairing supports naming a peak layer}. A
smaller $n = 60$ subset of these instances re-analyses to the same shape, with L36 at
$\ci{+0.067}{-0.025}{+0.167}$: non-significant there, significant at $n = 150$, and the point
estimate barely moved.

\paragraph{The destruction control, computed exactly.}
Equation~\ref{eq:distractor} is unbiased under uniform destruction. The destroyed answer distribution is
significantly non-uniform (passage-position patches, $\chi^2$ with $p = 2\times10^{-21}$; per-symbol
hit rates $0.061$ to $0.460$), and that is a property of the model rather than of our design, since the operator-table construction is indistinguishable from uniform at $p = 0.74$. With the realised
candidate set recorded per trial the residual is exact rather than bounded in expectation, and its
sign follows the donor pairing: $+0.0043$ at pairing 0 and $-0.0154$ at pairing 2, from mean destroyed
rates of $0.1121$ / $0.1100$ in the donor and distractor roles at the first and $0.1080$ / $0.1144$ at
the second. The worst single pair reaches $+0.1133$ and $+0.2171$ respectively, which supports an
adversarial bound in which \emph{every} pair is as bad as the worst observed:

\begin{center}
\small
\begin{tabular}{lcc}
\toprule
patch layer & survives the mean correction & survives ``every pair is the worst pair'' \\
\midrule
L39 & yes, both pairings & \textbf{yes} ($+0.1701$ / $+0.0962$) \\
L42 & yes, both pairings & \textbf{yes} ($+0.1700$ / $+0.0695$) \\
L45 & yes, both pairings & \textbf{yes} ($+0.0201$ / $+0.0229$) \\
\textbf{L36} & yes ($+0.1090$ / $+0.0821$) & \textbf{no} on either ($-0.0566$ / $-0.2071$) \\
L48--L57 & yes & no, \emph{and that is the point} \\
\bottomrule
\end{tabular}
\end{center}

\noindent L39--L45 therefore survive a bound stronger than anything the paper claims elsewhere, and
L36 is robust to the measured residual and to the pairing but not to the adversarial bound. The deep
cells fail it for a substantive reason rather than an estimator artifact: from L51 a residual patch
transplants the donor's near-final computation, which emits the donor's symbol without transporting a
latent variable. The contamination control shows the same thing from another angle. The predicted
counterfactual symbol coincides with the \emph{donor's own} gold answer in $8$ of $60$ pairs;
excluding those moves L39 from $0.433$ to $0.442$ and L48 from $0.483$ to $0.481$, but moves L57 from
$0.367$ to $0.288$, by far the largest move of any layer, and exactly where wholesale copying should
bite.

\paragraph{The span curve.}
Over $m$ patched query tokens, flips reach $\ci{+0.1250}{+0.0437}{+0.2125}$ at four positions with
distractors flat at $0.025$, then $+0.294$, $+0.313$, $+0.331$ and $+0.338$ at six, eight, ten and
twelve, while accuracy falls from $0.900$ to $0.550$. Four positions is the cleanest cell, significant
with accuracy still $0.800$; the widest span is partly overwriting. At one position the flip rate does
not clear zero but the graded margin does, $\ci{+0.550}{+0.271}{+0.831}$.

\paragraph{Survival, and what a passage patch cannot test.}
Reading the donor's value at a fixed later band (L46--L51) after a patch at L33--L45, at $n = 100$ and
span 1: L33 $\ci{+0.0048}{-0.0010}{+0.0113}$ (null), L36 $\ci{+0.0453}{+0.0326}{+0.0596}$, L39
$+0.0871$, L42 $+0.0877$, L45 $+0.0888$. The target's own value is suppressed monotonically over the
same range, $-0.0020$, $-0.0217$, $-0.0261$, $-0.0365$. Both hold in a second, smaller sample to the
digit ($-0.002$, $-0.017$, $-0.024$, $-0.035$; L36 at $51\%$ of asymptote against $45\%$) across two
spans and two sample sizes.

The repair account predicts that removing the source blocks survival. Patching the passage's last
twelve tokens from the same donor transports $+0.0007$ to $+0.0025$ to the answer position on its own,
and adds nothing on top of a query patch: at L39 and L42 the query-only, split-span and both-span
arms agree to four decimal places ($+0.0346$, $+0.0346$, $+0.0347$). The intervention is therefore too
weak to test the account, because the language is established across the whole passage rather than its
tail; blocking re-derivation properly means overwriting the passage entire, which is a different and
far more destructive experiment. We record it as a dead end whose cause is a property of the task, not
of the model.

\paragraph{Competition, in the per-trial data.}
Suppression tracks patch depth rather than donor load, and within a layer the trial-level correlation
between donor installed and target lost is \emph{positive} ($+0.156$ at L39 and $+0.553$ at L33),
where a capacity trade-off predicts negative. The narrower claim is what we make: deep substitution
displaces the incumbent value.

\paragraph{The mediating direction, both derivations.}
Since the readout is $W_U\,\mathrm{norm}(J_\ell h_\ell)$, the \emph{static} pre-normaliser lens vector
for token $z$ is $J_\ell^\top W_U[z]$, while the exact logit gradient is
$\nabla_h \lambda_z = J_\ell^\top\,DN(J_\ell h)^\top\,W_U[z]$, which for RMSNorm weights the
unembedding row by the gain \emph{and removes the component along the activation}. That second
property makes the exact gradient unusable as a mediator: the normaliser is scale-free, so the readout
does not change along the activation's own direction, and projecting the gradient out removes
identically none of the activation. We verified both derivations against automatic differentiation.
The static vector is the one with overlap to remove, so the correction is to the name rather than to
the intervention.

\paragraph{The mediation controls.}
The body table gives the absolute projection at L39. All of it, as ranges over three pairings:

\begin{center}
\footnotesize
\begin{tabular}{llcc}
\toprule
component & direction projected out & absolute projection & from $h_d - h_t$ \\
\midrule
\texttt{resid.L39} & the concept's own (gold) & $\mathbf{+0.113}$ to $\mathbf{+0.193}$ & $+0.138$ \\
 & gold, orthogonalised vs the other 19 & $\mathbf{+0.127}$ to $\mathbf{+0.213}$ & $+0.200$ \\
 & random, same activation norm & $-0.013$ to $+0.007$ & $-0.013$ \\
 & rivals, mean of $20$ & $-0.017$ to $+0.015$ & $+0.005$ \\
 & rivals, worst of $20$ & $+0.014$ to $+0.042$ & $+0.029$ \\
\midrule
\texttt{resid.L42} & the concept's own (gold) & $\mathbf{+0.080}$ to $\mathbf{+0.147}$ & $+0.087$ \\
 & gold, orthogonalised vs the other 19 & $\mathbf{+0.100}$ to $\mathbf{+0.127}$ & $+0.125$ \\
 & random, same activation norm & $-0.013$ to $+0.007$ & $+0.000$ \\
 & rivals, mean of $20$ & $-0.003$ to $+0.010$ & --- \\
 & rivals, worst of $20$ & $+0.014$ to $+0.042$ & --- \\
\bottomrule
\end{tabular}
\end{center}

\noindent Activation norms removed, so that a control which removes almost nothing is visible as
such: gold $13.63$, orthogonalised $9.50$, random $13.63$ by construction, rivals $9.6$. Dose response
for the gold direction at $0.25$/$0.50$/$0.75$/$1.00$ is $+0.062$, $+0.125$, $+0.150$, $+0.188$, and
flat at every dose for every control.

\textbf{The mediated \emph{share} is the least stable quantity in the paper and we do not quote it as
a point.} Computed consistently from observations (gold-removal cost over the full-arm donor-symbol
rate on the same paired trials, never from a stored summary string), it reads $45.8\%$, $33.3\%$ and
$43.9\%$ at L39 and $36.7\%$, $27.3\%$ and $18.5\%$ at L42 across the three pairings, against $50.0\%$
and $31.0\%$ at $n = 80$. It is a ratio of two pairing-sensitive quantities, which is
why it behaves worst of anything we measure; the smaller run's L42 value sits inside its range and its
L39 value above it, which is why the share is reported as a range and never as a point.

\section{Transport and the attention route, in detail}
\label{app:transport}

\paragraph{What the convergent literature says, verbatim.}
\citet{geva2023dissecting} block attention edges during factual recall and find the attribute
extracted to the prediction position by attention heads, with early MLPs instead enriching the subject
position. \citet{wang2023ioi} identify heads ``active at END'' that ``attend to previous names in the
sentence, and copy the names they attend to''. \citet{todd2024functionvectors} find that ``a small
number [of] attention heads transport a compact representation of the demonstrated task'', with the
strongest causal effects in middle layers, the same depth range as our window. And
\citet{gurnee2026workspace} report that ``a subset of attention heads relays J-space content between
positions, and ablating it selectively disrupts behaviors that depend on that relay''. None of these
manipulates task demand over a fixed context, and none separates installation depth from survival
depth, which is where our increment lies.

\begin{figure}[t]
\centering
\includegraphics[width=\textwidth]{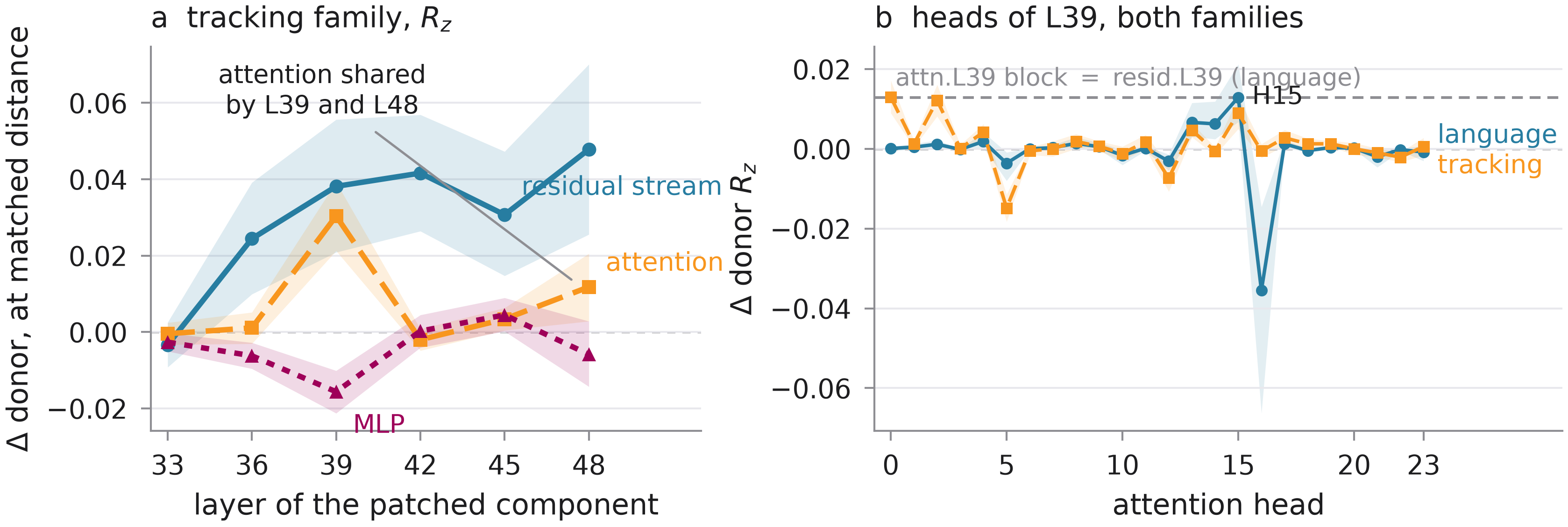}
\caption{The second task family, in $\Rz$ at $n = 140$, the readout the four-pairing table below
reports, named in the panel title because the language panels use $\Lz$. \textbf{(a)} The branch
asymmetry replicates: the stream and the attention output transport and the MLP at L39 opposes, but
the attention effect is shared between L39 and L48 rather than concentrated at one layer as it is on
language; L48 is a linear-attention layer. \textbf{(b)} The 24 heads of L39 in both families,
one donor pairing each; the language series is $n = 60$. On language one head matches both the block
and the whole stream to three decimal places; on tracking the same head participates at about a fifth
of its own family's stream, which four pairings confirm is nonzero (table below). Language depth
profile: Figure~\ref{fig:spine}a.}
\label{fig:gather}
\end{figure}

\paragraph{The second family, four donor pairings.}
At $n = 140$, distance 3--5 mean, $\Rz$, with Benjamini--Hochberg run inside each pairing:

\begin{center}
\small
\begin{tabular}{lccccl}
\toprule
component & p0 & p1 & p2 & p3 & BH \\
\midrule
\texttt{resid.L39} & $+0.0382$ & $+0.0346$ & $+0.0377$ & $+0.0505$ & \textbf{4/4} \\
\texttt{resid.L42} & $+0.0416$ & $+0.0348$ & $+0.0410$ & $+0.0411$ & \textbf{4/4} \\
\texttt{resid.L48} & $+0.0478$ & $+0.0526$ & $+0.0466$ & $+0.0619$ & \textbf{4/4} \\
\texttt{attn.L39} & $+0.0304$ & $+0.0186$ & $+0.0164$ & $+0.0271$ & \textbf{4/4} \\
\texttt{attn.L48} & $+0.0118$ & $+0.0204$ & $+0.0226$ & $+0.0094$ & \textbf{4/4} \\
\texttt{attn.L39.H15} & $+0.0090$ & $+0.0079$ & $+0.0058$ & $+0.0094$ & \textbf{4/4} \\
\texttt{attn.L39.H0} & $+0.0130$ & $+0.0121$ & $+0.0105$ & $+0.0116$ & \textbf{4/4} \\
\texttt{attn.L39.H5} & $-0.0148$ & $-0.0149$ & $-0.0136$ & $-0.0117$ & \textbf{4/4} negative \\
\texttt{mlp.L39} & $-0.0156$ & $-0.0113$ & $-0.0148$ & $-0.0117$ & \textbf{4/4} negative \\
\texttt{mlp.L42} & $+0.0003$ & $-0.0004$ & $+0.0035$ & $+0.0025$ & 0/4, null throughout \\
\texttt{attn.L39.H17} & $+0.0028$ & $+0.0020$ & $+0.0012$ & $+0.0036$ & 2/4, \emph{not stable} \\
\bottomrule
\end{tabular}
\end{center}

\noindent \texttt{attn.L48} and \texttt{attn.L39.H15} both clear zero under $\Rz$, $\Lz$ and $\Mz$
under every pairing, and survive
correction under $\Rz$ and $\Mz$ under every pairing and under $\Lz$ in three of four, with the failing pairing differing between the two cells, so their metric-dependence
is a property of one draw. \texttt{mlp.L42} is null in all four pairings, and
\texttt{attn.L39.H15} is nonzero on this family. Correction also promotes a cell we
would otherwise not mention: \texttt{attn.L39.H5} is significantly \emph{negative}, so L39 on tracking
is mildly opposed at one head while participating at others.

The head-level instability of \S\ref{sec:pairing} is worth quantifying here, because this sweep is
where we found it. Survivor counts per pairing are $21$ / $19$ / $15$ / $23$ of 42 under $\Rz$,
$23$ / $18$ / $16$ / $22$ under $\Lz$ and $26$ / $25$ / $21$ / $29$ under $\Mz$, but the four-way
intersections are only $14$, $13$ and $18$. Ten of the twelve cells significant in some pairings and
not others are \texttt{attn.L39.H*} at point estimates of $\pm 0.001$--$0.005$.

\paragraph{The head, and the source work's open question.}
\texttt{H13} ($+0.0099$) and \texttt{H14} ($+0.0093$) each carry roughly half of \texttt{H15}'s effect
on language, which is why we call the head sufficient \emph{for the block's transport effect} rather
than uniquely necessary.
The largest head effect at the layer is not \texttt{H15} but its neighbour \texttt{H16}, in the other
direction: $\ci{-0.0421}{-0.0724}{-0.0196}$, twice \texttt{H15}'s magnitude. So the layer contains a
head that installs the donor's value and one that opposes it, which is consistent with the gather
being selective routing and is a further reason the head is not the whole story. At
$n = 60$ and readout distance 4, \texttt{attn.L39.H15} gives $\ci{+0.0194}{+0.0098}{+0.0323}$, the
whole attention block $\ci{+0.0193}{+0.0098}{+0.0321}$ and the whole residual stream
$\ci{+0.0193}{+0.0097}{+0.0322}$; the same coincidence holds at distances 3 and 5. This run is
single-pairing.

\citet{gurnee2026workspace} note that the absence of verbalizable content in early layers may be real
\emph{or} may be lens degeneracy there, leaving open that a ``true workspace'' extends earlier. Our
behavioural counterfactual never touches the lens and is also null everywhere below L36. The
boundaries are not identical: the lens-derived measure already transports at L36 while the
behavioural one is weakest there. But both are far weaker below L36 than inside the window, and two
measures with disjoint failure modes agreeing there is evidence that the shallow weakness is
\textbf{real}. This is a claim about \emph{concentration}, not absence: the lens-derived measure does
find small significant shallow transport under $\Lz$, and what the two agree on is that the shallow
band carries far less.

\paragraph{The attention route.}
Measuring the route needs equal-length instructions across arms, which our released wording does not
have, so the family was regenerated with four 14-token arms sharing a nine-token identical tail; the
visibility effect survives the rewrite ($\ci{+0.1091}{+0.0965}{+0.1217}$ against
$\ci{+0.0891}{+0.0799}{+0.0983}$). On those records, one clean forward pass each and no patching,
comparing L39 against the 72 heads of L15, L27 and L51 at $n = 200$: L39's \emph{largest} head against
the controls' \emph{median} gives $4.5\times$, median with median $1.90\times$, maximum with maximum
$1.16\times$. We report the middle one. A permutation that shuffles head labels among all 96 heads
makes a randomly chosen layer score $3.46\times$ on the asymmetric max-over-median statistic, which is
why we neither quote that statistic nor attach a $p$-value to any of them: the inferential unit for a
claim about a layer is the layer, and we measured four. Correction is likewise uninformative here,
since 94 of 96 heads survive at $q = 0.05$. \texttt{H15} shifts by
$\ci{+0.1546}{+0.1486}{+0.1603}$, the $91.7$th percentile of the 72 control heads, and is positive
against the same head index at all three control layers. Twenty-three of L39's 24 heads shift toward
the passage, which is not distinctive: L27 gives 21 of 24 and L51 gives 20 of 24. On the single final
token, identical across arms by construction, a control layer leads instead, so the effect is a
property of the nine-token span and we report it as such.

\section{Cross-architecture screening, and the checkpoints that failed it}
\label{app:across}

\begin{figure}[t]
\centering
\includegraphics[width=\textwidth]{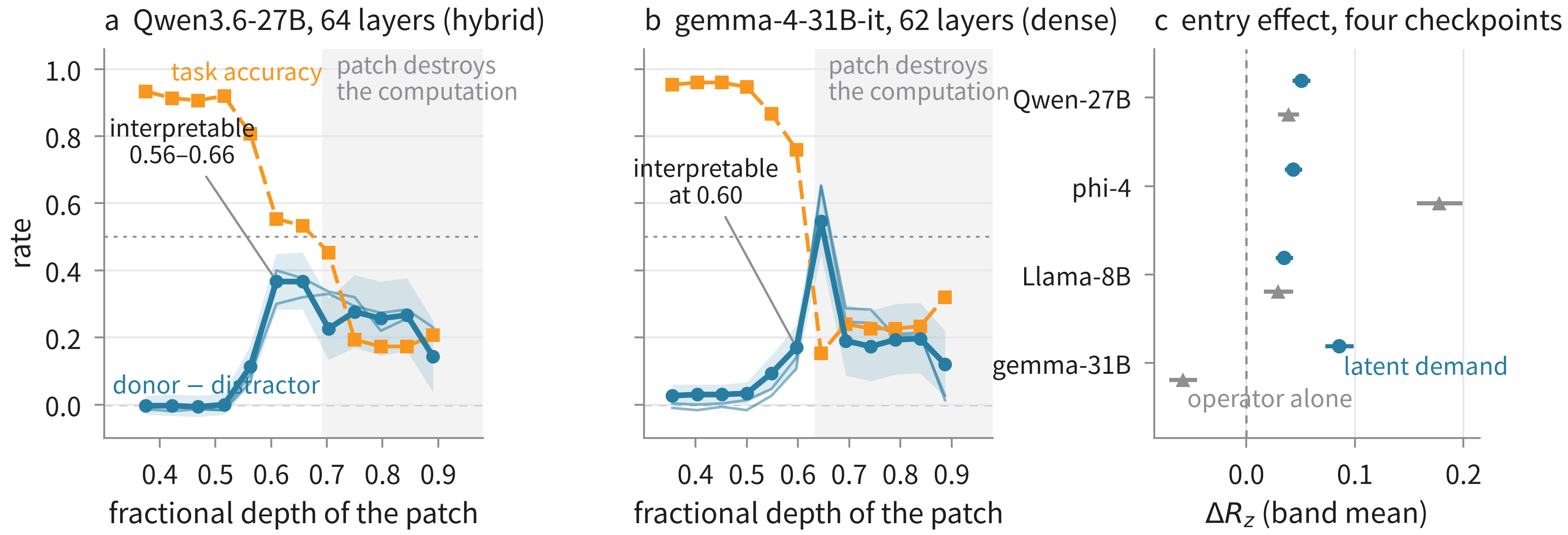}
\caption{What travels across checkpoints. \textbf{(a, b)} The distractor-controlled counterfactual
against fractional patch depth at $n = 150$ pairs per pairing, on the same axes, with all three donor
pairings drawn (the thick line carries the bootstrap band) and task accuracy dashed. Where accuracy
falls below $0.5$ the patch has replaced the late computation, and the shading marks that a
donor-symbol rate there is not evidence of transport. \textbf{(c)} The entry effect with the operator
held fixed (\emph{flexible} $-$ \emph{supplied}) at $n = 200$ instances per checkpoint is positive on
all four; the operator's own contribution (\emph{supplied} $-$ \emph{control}) is not stable in sign.}
\label{fig:across}
\end{figure}

Five checkpoints carry a number in this paper and two more were measured and excluded; the screens
table below covers the remaining four, Qwen3.5-9B being a replication of the entry effect
alone, on its own lens. Nothing here is
inferred from a model family: every exclusion is a measurement.

\paragraph{The screens.}
A candidate must (i) tokenize all 20 language names and enough operator symbols in \emph{continuation}
form, with the leading space forming a distinct token; (ii) pass the homogeneity check of
Appendix~\ref{app:gauge} if a lens-derived claim is to rest on it; (iii) be dense, since ``the
instruction changed'' against ``the router picked different experts'' is exactly this paper's confound;
and (iv) for the depth sweep only, solve the flexible arm near ceiling. One candidate fails (i):

\begin{center}
\small
\begin{tabular}{lccc}
\toprule
checkpoint & languages single-token & operator symbols & \texttt{" X"} $\ne$ \texttt{"X"} \\
\midrule
Qwen3.6-27B & 20/20 & 9/18 & yes, 4 of 4 probes \\
phi-4 & 20/20 & 9/18 & yes, 4 of 4 \\
Llama-3.1-8B-Instruct & 20/20 & 9/18 & yes, 4 of 4 \\
gemma-4-31B-it & 20/20 & 9/18 & yes, 4 of 4 \\
\midrule
Mistral-Small-3.2-24B & \textbf{7/20} & 18/18 & \textbf{no, 0 of 4} \\
\bottomrule
\end{tabular}
\end{center}

\noindent The Mistral tokenizer strips the leading space: \texttt{" Spanish"} and \texttt{"Spanish"}
are the same id, and \texttt{" "} encodes to the empty list, so invariant 4 does not hold there and
``score the continuation form'' is not the same operation as on the other four. A cross-model claim
would be comparing two different measurements. Separately, one Gemma-3 checkpoint fails (ii) with a gated MLP homogeneous of degree $1.76$--$1.93$, and its published lens underflows in fp16, which is why gemma-4 rather than gemma-3
carries the cross-architecture window. We re-checked gemma-4 against the same criteria rather than
inheriting the family's exclusion: it is dense, it passes the tokenizer screen, and we fitted its lens
ourselves.

\paragraph{Per-arm accuracy, which is what makes the $2\times2$ non-comparable.}
Re-pooled from each run's per-trial file rather than from a stored summary:

\begin{center}
\footnotesize
\begin{tabular}{lccccc|c}
\toprule
checkpoint & automatic & control & supplied & report & \textbf{flexible} & operator share \\
\midrule
Qwen3.6-27B & $0.615$ & $0.940$ & $\mathbf{1.000}$ & $0.980$ & $0.940$ & $43\%$ \\
Llama-3.1-8B-Instruct & $0.815$ & $0.680$ & $\mathbf{1.000}$ & $0.995$ & $0.610$ & $46\%$ \\
phi-4 & $0.805$ & $0.990$ & $0.860$ & $0.970$ & $\mathbf{0.515}$ & $81\%$ \\
gemma-4-31B-it & $0.360$ & $\mathbf{1.000}$ & $\mathbf{1.000}$ & $0.845$ & $0.970$ & --- \\
\bottomrule
\end{tabular}
\end{center}

\noindent A ceiling or floor arm is a design failure rather than a finding, and both failures are
present. Qwen and Llama measure the operator-only contrast against an arm at exactly $1.000$, so it has
no behavioural variance, and those are the two whose shares agree, to three points. phi-4 escapes
that and has the opposite problem, a \emph{flexible} arm failing half the time. gemma-4 has both arms
of the operator contrast pinned at $1.000$ and a significantly \emph{negative} operator effect,
$\ci{-0.0585}{-0.0691}{-0.0476}$, so its share is not a percentage of anything and we print no number
rather than a large absurd one. Also: Llama's
\emph{control} arm at $0.680$ sits \emph{below} its automatic arm at $0.815$, the only checkpoint where
that happens, so its format-matched control is not the easy arm it is elsewhere.

What is comparable across checkpoints is the operator-controlled contrast, and it is positive on all
four:

\begin{center}
\footnotesize
\begin{tabular}{lcc}
\toprule
checkpoint & latent demand (flexible $-$ supplied) & operator alone (supplied $-$ control) \\
\midrule
Qwen3.6-27B & $\ci{+0.0504}{+0.0445}{+0.0567}$ & $\ci{+0.0387}{+0.0310}{+0.0466}$ \\
phi-4 & $\ci{+0.0429}{+0.0371}{+0.0493}$ & $\ci{+0.1778}{+0.1590}{+0.1970}$ \\
Llama-3.1-8B & $\ci{+0.0346}{+0.0288}{+0.0408}$ & $\ci{+0.0292}{+0.0178}{+0.0409}$ \\
gemma-4-31B & $\ci{+0.0855}{+0.0743}{+0.0967}$ & $\ci{-0.0585}{-0.0691}{-0.0476}$ \\
\bottomrule
\end{tabular}
\end{center}

\noindent Interactions on the same runs are $\ci{-0.0487}{-0.0571}{-0.0405}$,
$\ci{-0.1864}{-0.2059}{-0.1672}$, $\ci{-0.0499}{-0.0624}{-0.0379}$ and
$\ci{+0.0046}{-0.0075}{+0.0169}$ in that order: sub-additive on three and null on gemma-4. Note that
gemma-4 is also the best accuracy-matched of the four on the latent-demand contrast, $0.970$ against
$1.000$. It gives the largest effect \emph{in percentile rank} and the smallest under the
non-saturating readout ($\ci{+0.1679}{+0.1441}{+0.1938}$ against Qwen's
$\ci{+1.3134}{+1.2548}{+1.3718}$, phi-4's $\ci{+1.3921}{+1.3155}{+1.4668}$ and Llama's
$\ci{+1.5438}{+1.4644}{+1.6223}$), so the ordering across checkpoints is a fact about the measure.
All four clear zero under both. An accuracy-matched \emph{supplied} arm on a second checkpoint
would settle the decomposition properly; it is not run and is not cheap.

\paragraph{The window on gemma-4-31B-it.}
Sixty-two layers, dense, clean flexible accuracy $0.970$. The counterfactual is lens-free, so nothing
here inherits from a lens artifact. Three donor pairings, $n = 150$:

\begin{center}
\footnotesize
\setlength{\tabcolsep}{3.5pt}
\begin{tabular}{llcccc}
\toprule
patch layer & frac.\ depth & pairing 0 & pairing 1 & pairing 2 & accuracy \\
\midrule
L22--L31 & $0.35$--$0.50$ & $+0.027$ & $-0.017$ & $+0.000$ & $0.93$--$0.96$ \\
L34 & $0.55$ & $+0.093^{*}$ & $+0.027$ & $+0.047^{*}$ & $0.87$ \\
\textbf{L37} & $\mathbf{0.60}$ & $\mathbf{+0.170^{*}}$ & $\mathbf{+0.107^{*}}$ & $\mathbf{+0.140^{*}}$ & $0.74$--$0.77$ \\
L40 & $0.65$ & $+0.547^{*}$ & $+0.650^{*}$ & $+0.653^{*}$ & $0.09$--$0.15$ \\
L43--L55 & $0.69$--$0.89$ & $+0.120$ to $+0.197$ & $+0.013$ to $+0.247$ & $+0.027$ to $+0.287$ & $0.17$--$0.27$ \\
\bottomrule
\end{tabular}
\end{center}

\noindent Starred cells exclude zero. L37's intervals are $\ci{+0.170}{+0.103}{+0.237}$,
$\ci{+0.107}{+0.047}{+0.170}$ and $\ci{+0.140}{+0.073}{+0.210}$; L34's least favourable is
$\ci{+0.027}{-0.017}{+0.070}$, which is why it is two of three. \textbf{L37 is the only cell
significant under all three pairings at an accuracy that still permits a reading.} The cells from L40
upward carry the largest effects in this paper and none of them is transport.

Splitting that layer by component kind, \textbf{at one donor pairing only}: \texttt{resid.L37}
$\ci{+0.1700}{+0.1033}{+0.2367}$ at accuracy $0.760$, \texttt{attn.L37}
$\ci{+0.0200}{-0.0067}{+0.0500}$ at $0.960$, and \texttt{mlp.L37} $\ci{+0.0200}{-0.0033}{+0.0500}$ at
$0.960$. So no component reproduces the stream's behavioural effect here either, which replicates the
joint-component negative of \S\ref{sec:window} on a second architecture. It is \emph{not} a replication
of \S\ref{sec:illusion}: that result is about a readout shift without a behavioural effect, and we
have no readout measurement on this checkpoint. \texttt{attn.L40} reaches
$\ci{+0.0400}{+0.0067}{+0.0767}$ at accuracy $0.940$, the one significant component-level behavioural
cell here.

\paragraph{The checkpoint the design cannot use.}
Llama-3.1-8B-Instruct clears every screen and still cannot resolve the window, because it solves the
flexible arm at $0.610$. At $n = 250$, span 12: L12 (fractional depth $0.38$) is null at accuracy
$0.396$; L14 ($0.44$) reaches $\ci{+0.150}{+0.068}{+0.232}$ but at accuracy $0.208$; L16
$\ci{+0.174}{+0.088}{+0.260}$ at $0.172$; L18 and L20 are significant at $0.19$--$0.21$; L22 and above
are null. At span 1 the pattern repeats with slightly more headroom: L14
$\ci{+0.136}{+0.060}{+0.212}$ at accuracy $0.336$, L12 null at $0.500$. \textbf{No cell is both
significant and at an accuracy comparable to the interpretable cells on the other two checkpoints},
where the analogous figures are $0.53$--$0.81$. The distractor rate is also high throughout
($0.34$--$0.52$), which is what a patch does when the clean computation has little margin. Its entry
effect replicates normally, on the one published lens in our set that ships both a configuration file
and a convergence trace, and that trace shows the fitter stopping on a convergence delta at $461$ of
a requested $1000$ prompts, which is worth noting because the flagship 27B artifact ships neither file
while its name records $n = 1000$.

\section{Supporting figures and tables}
\label{app:tables}

\begin{figure}[t]
\centering
\includegraphics[width=\textwidth]{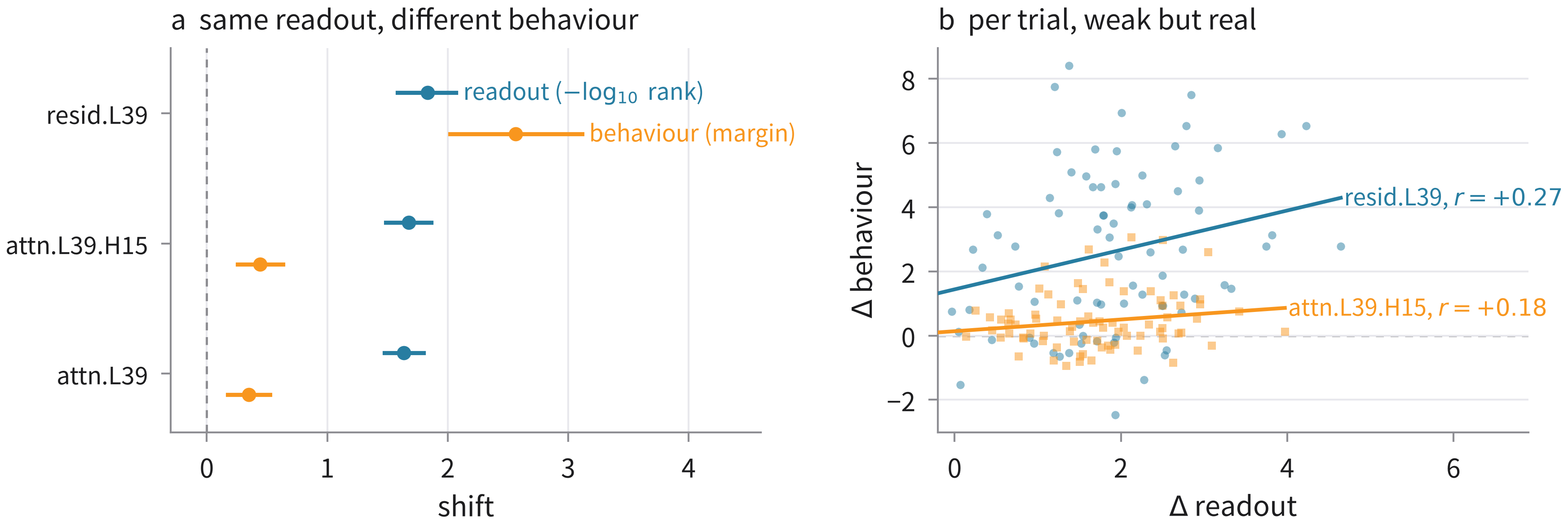}
\caption{A readout shift is not a calibrated measure of use, on the same $80$ trials throughout, one
donor pairing. \textbf{(a)} Three components at L39 shift the donor concept's readout to within
$12\%$ of one another while their effect on behaviour differs $7.4\times$. \textbf{(b)} Per trial the
two are related but weakly: uncalibrated rather than uninformative.}
\label{fig:calibration}
\end{figure}

\begin{figure}[t]
\centering
\includegraphics[width=\textwidth]{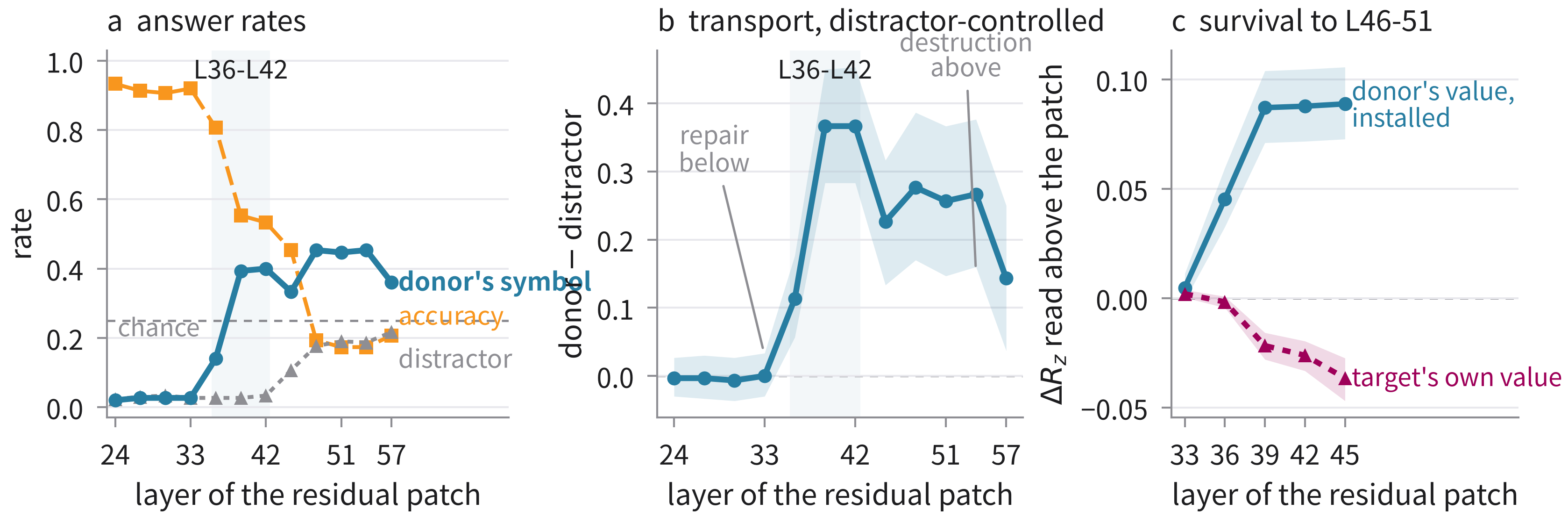}
\caption{The behavioural window and both of its edges. \textbf{(a)} Rates against patch depth at
$n = 150$, one pairing: the donor's predicted symbol, the matched per-distractor rate, and task
accuracy. From L48 the distractor rate approaches chance while accuracy is below $0.2$, which is
destruction rather than transport. \textbf{(b)} The distractor-controlled effect; the shaded band
marks the cells where accuracy still permits a reading. \textbf{(c)} The lower edge is a survival
failure, at $n = 100$ and span 1: a value installed at L33 is gone by L46, and the target's own value
is progressively suppressed as the donor's is installed.}
\label{fig:window}
\end{figure}

\begin{figure}[t]
\centering
\includegraphics[width=\textwidth]{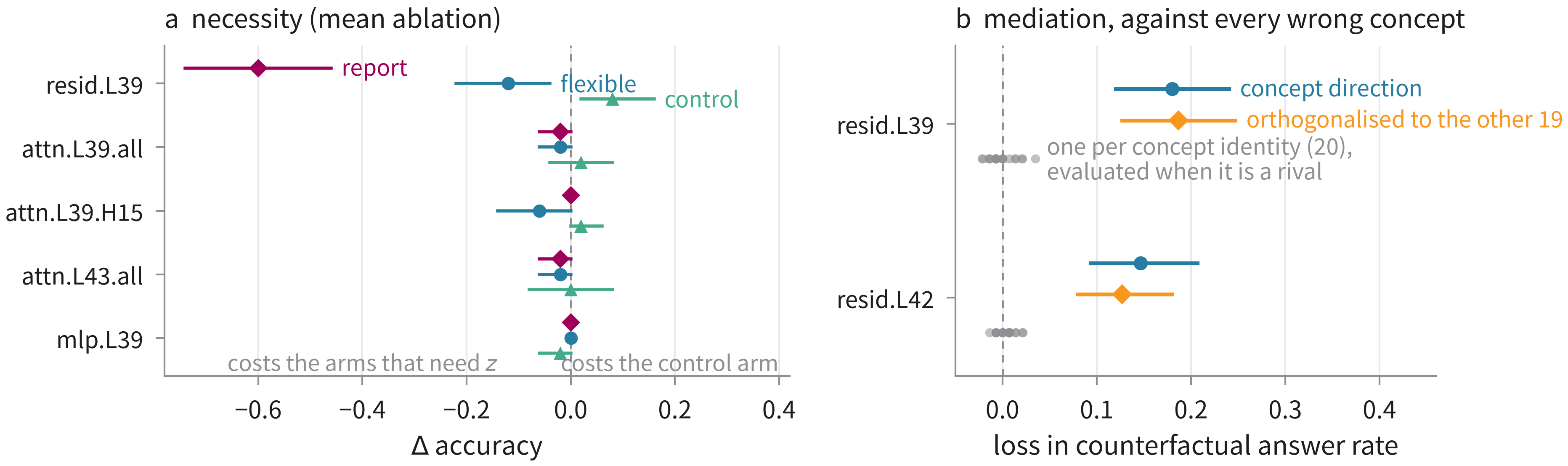}
\caption{\textbf{(a)} Necessity by \emph{leave-one-out mean} ablation, $50$ instances per arm. A
component is selectively necessary when it costs the arms that need $z$ and not the format-matched
control; only \texttt{resid.L39} clears zero under this ablation, and \texttt{mlp.L39} is the one cell
pointing the other way. That the control arm is measurable at all is shown by the harsher zero
ablation in Table~\ref{tab:ablate}, not by this panel.
\textbf{(b)} Mediation at $n = 150$, one pairing: the loss in counterfactual answer rate from
projecting out the concept's static lens vector, and from the same vector orthogonalised against the
other nineteen concepts, against a null drawn one point per wrong concept. Paired per instance
throughout, so a rival that happened to help would appear left of zero. The null is plotted as one
estimate per concept identity rather than as one interval, because a single random control cannot be
told from a lucky draw.}
\label{fig:mechanism}
\end{figure}

\begin{figure}[t]
\centering
\includegraphics[width=\textwidth]{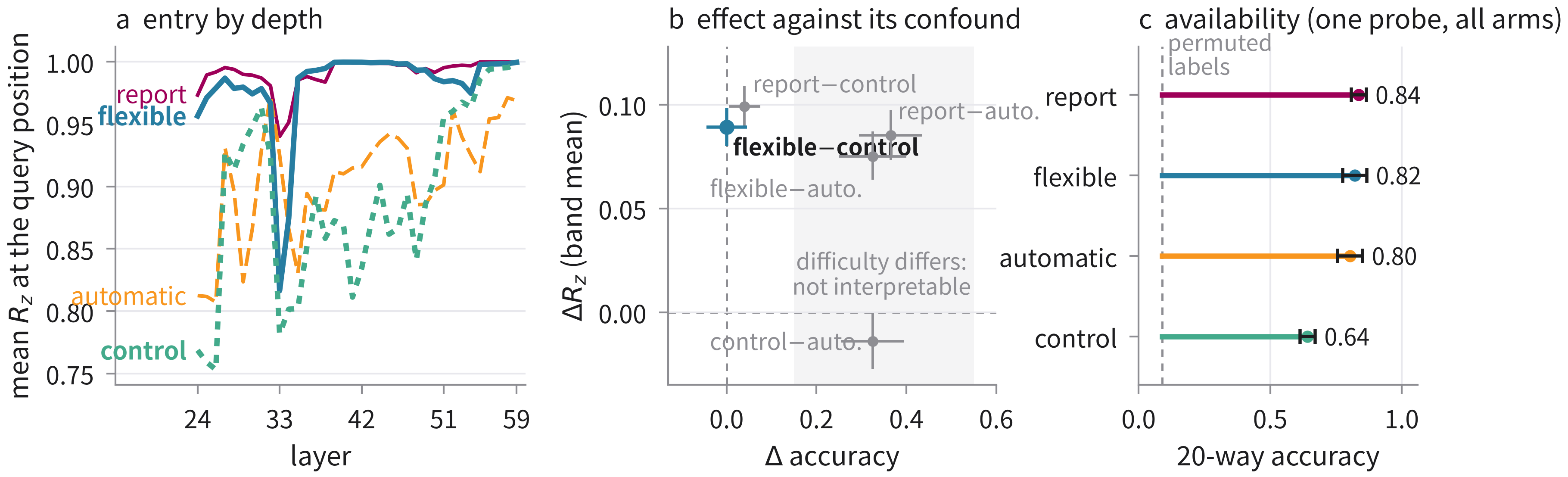}
\caption{Visibility dissociates from availability, $n = 200$ semantic instances throughout.
\textbf{(a)} Mean $\Rz$ of the gold concept at the
query position by layer and condition. The L33 dip appears in all four arms but is about three times
deeper in flexible and control than in report and automatic, so it cancels within either pair and not
across them. \textbf{(b)} Each contrast's visibility effect against the accuracy difference it must
not be; the primary contrast sits at $\Delta$accuracy $=0$, and shading marks where difficulty differs
enough that the effect is not interpretable. \textbf{(c)} A single linear probe trained jointly across
all four arms decodes $z$ in every one. \textbf{This panel plots the layer-selected form}, whose
diagonal runs $0.64$ (control) to $0.84$ (report); the nested figures quoted in \S\ref{sec:entry},
$0.575$ to $0.808$, choose the layer on an inner split of the training instances only and are the ones
the claim rests on. Bars are the mean over five instance-disjoint splits and whiskers their standard
deviation; because the layer is selected, the reference line is the $0.090$ reached by permuted labels
under the same selection rule, not the nominal $0.05$.}
\label{fig:entry}
\end{figure}

\begin{table}[t]
\centering
\footnotesize
\setlength{\tabcolsep}{4pt}
\begin{tabular}{llrl}
\toprule
role & artifact & $n$ & unit \\
\midrule
\multicolumn{4}{l}{\emph{selection: small by design, frozen before everything below}} \\
coarse screen (chose L39) & \texttt{coarse20} & 20 & pairs \\
head decomposition (chose \texttt{H15}) & \texttt{heads\_full} & 25 & pairs \\
bidirectional screen & \texttt{coarse60} & 60 & pairs \\
\midrule
\multicolumn{4}{l}{\emph{entry and availability}} \\
Stage 1, four arms & \texttt{s1v2} & 200 & instances \\
Stage 1, five arms ($2\times2$) & \texttt{twobytwo\_n200} & 200 & instances \\
Stage 1, phi-4 / Llama / gemma-4 & \texttt{*\_stage1} & 200 & instances \\
linear probe & \texttt{probe2} & 200 & instances \\
\midrule
\multicolumn{4}{l}{\emph{transport}} \\
matched-distance sweep, language ($\times4$) & \texttt{n120\_grid\_s*} & 120 & pairs \\
matched-distance sweep, control arm & \texttt{n120\_grid\_control} & 120 & pairs \\
head sweep, language & \texttt{A\_heads\_n60} & 60 & pairs \\
matched-distance sweep, tracking ($\times4$) & \texttt{F\_track\_gather\_n140*} & 140 & pairs \\
attention route, matched lengths & \texttt{G\_route\_len} & 200 & instances \\
\midrule
\multicolumn{4}{l}{\emph{behaviour}} \\
counterfactual, depth ($\times3$) & \texttt{cf\_resid\_depth\_n150*} & 150 & pairs \\
counterfactual, component groups & \texttt{cf\_groups} & 60 & pairs \\
counterfactual, span curve $m = 1 \ldots 12$ & \texttt{span\_m*} & 80 & pairs \\
readout against behaviour, same trials & \texttt{trial\_level} & 80 & pairs \\
survival of an installed value & \texttt{repair\_span1\_n100} & 100 & pairs \\
repair source, 4 position modes & \texttt{repair\_source\_v2\_n100} & 100 & pairs \\
necessity, leave-one-out ablation & \texttt{loo\_necessity} & 50 & instances/arm \\
mediation ($\times3$) & \texttt{wide\_n150*} & 150 & pairs \\
\midrule
\multicolumn{4}{l}{\emph{cross-architecture}} \\
depth sweep, gemma-4 ($\times3$) & \texttt{cf\_depth\_gemma4\_n150*} & 150 & pairs \\
component kind, gemma-4 & \texttt{cf\_kind\_gemma4} & 150 & pairs \\
depth sweep, Llama, spans 12 and 1 & \texttt{cf\_depth\_llama*} & 250 & pairs \\
homogeneity screens & \texttt{gauge\_s2\_k*} & --- & layers \\
\bottomrule
\end{tabular}
\label{tab:runs}
\caption{Every run in the paper with the sample it used, read from the \texttt{args} block of the
artifact it names so the table cannot drift from what was run. Artifact names are given without their directory (one per experiment family under the data root) and without the model slug each file
carries as a suffix. Instance selection is deterministic (sorted by identifier, then truncated), so a
rerun sees the same instances. Where a range over donor pairings is reported the seeds are $0$--$3$;
otherwise seed $0$. The probe averages five split seeds.}
\end{table}

\paragraph{One component carries the demand setting in both directions.}
The bidirectional screen is exploratory and reported uncorrected over its 52 components, so it
supports no significance claim of its own. In it exactly one component shows an effect that reverses
with the direction of substitution: \texttt{attn.L39}. Writing the flexible arm's L39 attention output
into the automatic run raises visibility by $\ci{+0.0517}{+0.0336}{+0.0715}$; writing the automatic
arm's into the flexible run lowers it by $\ci{-0.0270}{-0.0369}{-0.0188}$. No other component has both
directions excluding zero with opposite signs. We report it because it is the strongest evidence in our
own data for the routing-gate reading of \S\ref{sec:limitations}, and because it is in the released
artifacts either way.

\paragraph{Two uninterpretable arms, and the guard they produced.}
At three swaps the tracking family is uninterpretable: the report arm scores $0.125$ against its own
$0.250$ chance floor while control scores $0.915$, so $\Delta$accuracy is $-0.790$ for
report $-$ control. The run nonetheless returned an entry-effect verdict, because at that point the
guard compared accuracy against a fixed constant and had no notion of a per-arm chance floor. This is
the run that made us add one, and the stored artifact still carries the pre-guard verdict. At one swap
the family works ($0.975$ / $0.940$) and the visibility effect is twice the language family's. A family
interpretable at only one difficulty setting is a fact about the family, and we report it rather than
only the setting that worked. On Qwen3.5-9B the guard then did its job unaided: the usable contrast is
report rather than flexible, because the flexible arm reaches $0.420$ against control's $0.985$, and the
guard refuses flexible $-$ control as confounded at $\Delta$accuracy $= -0.565$ despite a nominal
$\Delta\Rz = +0.2602$, which, had it been reported, would have been the largest visibility effect in
the paper and a difficulty difference. Chance floors are per arm and differ within a family: on
tracking at one swap they are $0.25$ for flexible and report and $0.50$ for automatic and control, and
the automatic arm sits at $0.510$ against its $0.500$ floor, which is why no tracking result here uses
an automatic baseline.

\begin{table}[h]
\centering
\small
\begin{tabular}{lcc}
\toprule
train $\rightarrow$ test & raw & per-arm centred \\
\midrule
report $\rightarrow$ flexible & $0.625$ & $\mathbf{0.825}$ \\
flexible $\rightarrow$ report & $0.525$ & $\mathbf{0.750}$ \\
control $\rightarrow$ automatic & $0.213$ & $\mathbf{0.675}$ \\
control $\rightarrow$ flexible & $0.175$ & $\mathbf{0.625}$ \\
control $\rightarrow$ report & $0.163$ & $\mathbf{0.613}$ \\
report $\rightarrow$ automatic & $0.175$ & $\mathbf{0.600}$ \\
\bottomrule
\end{tabular}
\label{tab:transfer}
\caption{Cross-arm probe transfer, $n = 200$ instances, 20 classes, chance $0.05$, split by semantic
instance. Raw transfer
is weak because the arms differ in prompt format and occupy different regions of activation space;
removing each arm's own mean lifts transfer to the level of the within-arm diagonal ($0.64$--$0.84$;
Table~\ref{tab:transfer}).
Both columns are reported because the centring was added after seeing the raw numbers.}
\end{table}

\begin{table}[h]
\centering
\footnotesize
\begin{tabular}{lrrrrrr}
\toprule
 & \multicolumn{3}{c}{$\Delta\Rz$} & \multicolumn{3}{c}{$\Delta\Lz$} \\
\cmidrule(lr){2-4}\cmidrule(lr){5-7}
patch layer & stream & attention & MLP & stream & attention & MLP \\
\midrule
L12 & $+0.0028$ & $-0.0003$ & $+0.0004$ & $+0.020^{1}$ & $+0.004$ & $+0.001$ \\
L15 & $+0.0071^{1}$ & $+0.0009^{1}$ & $+0.0020$ & $+0.023^{1}$ & $+0.001$ & $+0.006$ \\
L18 & $+0.0051^{1}$ & $-0.0003$ & $+0.0001$ & $\mathbf{+0.051}^{4}$ & $+0.002$ & $+0.011^{1}$ \\
L21 & $+0.0031^{2}$ & $-0.0001$ & $-0.0002$ & $\mathbf{+0.063}^{4}$ & $+0.002$ & $-0.001$ \\
L24 & $+0.0003$ & $+0.0002^{1}$ & $+0.0001$ & $+0.030^{1}$ & $+0.004^{1}$ & $+0.002$ \\
L27 & $-0.0004$ & $+0.0005^{1}$ & $-0.0003$ & $+0.001$ & $+0.007^{1}$ & $-0.005^{1}$ \\
L30 & $-0.0069$ & $-0.0015^{1}$ & $-0.0002$ & $-0.026$ & $-0.015^{3}$ & $-0.005$ \\
L33 & $+0.0011^{1}$ & $+0.0019^{2}$ & $-0.0009^{2}$ & $-0.021^{1}$ & $+0.010^{1}$ & $-0.009^{1}$ \\
\midrule
L36 & $\mathbf{+0.0073}^{4}$ & $\mathbf{+0.0026}^{4}$ & $-0.0010^{2}$ & $\mathbf{+0.437}^{4}$ & $\mathbf{+0.104}^{4}$ & $\mathbf{-0.044}^{4}$ \\
\textbf{L39} & $\mathbf{+0.0137}^{4}$ & $\mathbf{+0.0135}^{4}$ & $\mathbf{-0.0301}^{4}$ & $\mathbf{+1.908}^{4}$ & $\mathbf{+1.618}^{4}$ & $\mathbf{-0.400}^{4}$ \\
L42 & $\mathbf{+0.0228}^{4}$ & $-0.0011^{1}$ & $\mathbf{-0.0105}^{4}$ & $\mathbf{+1.596}^{4}$ & $-0.011^{2}$ & $\mathbf{-0.086}^{4}$ \\
L45 & $\mathbf{+0.0965}^{4}$ & $+0.0008$ & $-0.0040^{2}$ & $\mathbf{+1.062}^{4}$ & $+0.006$ & $+0.000$ \\
L48 & $\mathbf{+0.1056}^{4}$ & $\mathbf{+0.0141}^{4}$ & $-0.0023$ & $\mathbf{+0.908}^{4}$ & $\mathbf{+0.064}^{4}$ & $-0.004$ \\
\bottomrule
\end{tabular}
\caption{Language family, transport at matched readout distance (3--5 layers above the patch),
$n = 120$ pairs, \textbf{four donor pairings}. Each cell is the mean over the four, and the
superscript counts how many of them have an interval excluding zero; bold is all four. Both rank
readouts are shown because they rank the layers differently: $\Rz$ makes L48 the largest cell, while $\Lz$, which does not compress the top of the vocabulary, puts the peak at L39 and reduces
\texttt{attn.L48} to a twenty-fifth of \texttt{attn.L39}. The window is where stability lives: every
stream cell from L36 to L48 survives all four pairings, while below it only \texttt{resid.L18} and
\texttt{resid.L21} do, and under $\Rz$ no shallow cell does at all. Two cells at L33 hold in only one or two
pairings of four: a significantly negative \texttt{mlp.L33} and a positive \texttt{attn.L33}.
Distances 3, 4 and 5 are averaged per
instance before pooling. Benjamini--Hochberg at $q = 0.05$ is run inside each pairing over this sweep's 39
cells and changes no cell here. Plotted in Figure~\ref{fig:spine}a ($\Lz$).}
\label{tab:transport}
\end{table}

\begin{table}[h]
\centering
\small
\begin{tabular}{lcc}
\toprule
patched group & donor $-$ distractor & accuracy after \\
\midrule
\texttt{resid.L42} & $\ci{+0.4250}{+0.2833}{+0.5667}$ & $0.450$ \\
attention, L36 $+$ L39 $+$ L42 & $\ci{-0.0333}{-0.0833}{+0.0167}$ & $0.883$ \\
MLP, L36 $+$ L39 $+$ L42 & $\ci{-0.0250}{-0.0583}{+0.0000}$ & $0.950$ \\
attention $+$ MLP, L36--L42, all 14 components & $\ci{+0.0250}{-0.0167}{+0.0833}$ & $0.950$ \\
\bottomrule
\end{tabular}
\label{tab:joint}
\caption{Joint component groups do not reproduce the stream's behavioural effect. Clean accuracy
$0.967$; $n = 60$ pairs, twelve query positions, one run and one donor pairing, so that the stream
reference and the groups are measured on the same instances.}
\end{table}

\begin{table}[h]
\centering
\scriptsize
\setlength{\tabcolsep}{3pt}
\begin{tabular}{llccc}
\toprule
ablation & component & report & flexible & control \\
\midrule
\multirow{5}{*}{\shortstack[l]{leave-one-out\\mean}}
 & \textbf{\texttt{resid.L39}} & $\mathbf{\ci{-0.600}{-0.740}{-0.460}}$
   & $\mathbf{\ci{-0.120}{-0.220}{-0.040}}$ & $\mathbf{\ci{+0.080}{+0.020}{+0.160}}$ \\
 & \texttt{attn.L39} & $\ci{-0.020}{-0.060}{+0.000}$ & $\ci{-0.020}{-0.060}{+0.000}$
   & $\ci{+0.020}{-0.040}{+0.080}$ \\
 & \texttt{attn.L39.H15} & $\ci{+0.000}{+0.000}{+0.000}$ & $\ci{-0.060}{-0.140}{+0.000}$
   & $\ci{+0.020}{+0.000}{+0.060}$ \\
 & \texttt{attn.L43} & $\ci{-0.020}{-0.060}{+0.000}$ & $\ci{-0.020}{-0.060}{+0.000}$
   & $\ci{+0.000}{-0.080}{+0.080}$ \\
 & \texttt{mlp.L39} & $\ci{+0.000}{+0.000}{+0.000}$ & $\ci{+0.000}{+0.000}{+0.000}$
   & $\ci{-0.020}{-0.060}{+0.000}$ \\
\midrule
\multirow{5}{*}{zero}
 & \texttt{resid.L39} & $\ci{-0.440}{-0.580}{-0.300}$ & $\ci{-0.640}{-0.760}{-0.500}$
   & $\ci{-0.500}{-0.660}{-0.340}$ \\
 & \texttt{attn.L39} & $\ci{+0.000}{+0.000}{+0.000}$ & $\ci{-0.080}{-0.160}{-0.020}$
   & $\ci{+0.060}{+0.000}{+0.140}$ \\
 & \texttt{attn.L39.H15} & $\ci{+0.000}{+0.000}{+0.000}$ & $\ci{-0.080}{-0.160}{-0.020}$
   & $\ci{+0.060}{+0.000}{+0.140}$ \\
 & \texttt{attn.L43} & $\ci{+0.000}{-0.060}{+0.060}$ & $\ci{-0.060}{-0.140}{+0.000}$
   & $\ci{+0.020}{-0.060}{+0.100}$ \\
 & \textbf{\texttt{mlp.L39}} & $\ci{+0.000}{+0.000}{+0.000}$ & $\ci{+0.000}{+0.000}{+0.000}$
   & $\mathbf{\ci{-0.220}{-0.340}{-0.120}}$ \\
\bottomrule
\end{tabular}
\caption{Ablation cost in accuracy by arm: $\Delta$ against the unablated run on the same $50$
instances per arm, twelve query positions, clean accuracies report $0.980$, flexible $0.960$, control
$0.920$. Under the leave-one-out mean ablation \S\ref{sec:mechanism} uses, \texttt{resid.L39} is the
only component with any interval excluding zero, and it is selective: it costs both arms that need $z$
and \emph{helps} the format-matched control. \texttt{attn.L39.H15} points the same way in the flexible
arm but does not clear zero there. The zero rows are the positive control for those nulls: zeroing
\texttt{mlp.L39} costs \emph{control} alone, which is what shows the control arm is not sitting at a
measurement floor. Zeroing \texttt{resid.L39} damages every arm and so distinguishes nothing; it is
printed to make that visible. Plotted, mean rows only, in Figure~\ref{fig:mechanism}a.}
\label{tab:ablate}
\end{table}

\paragraph{The largest MLP cell outside the window.}
On language, \texttt{mlp.L15} is the largest positive MLP point estimate outside the transport window
at matched distance, $+0.0020$ under $\Rz$ and $+0.0062$ under $\Lz$ as the mean over four donor
pairings, at a depth where the residual stream itself does not transport significantly. No pairing
resolves it from zero (Table~\ref{tab:transport}), so we report it as the least convenient cell we
found rather than as a counterexample, and it is why \S\ref{sec:gather} scopes the MLP claim to the
window rather than asserting it of MLPs in general.

\section{A worked example}
\label{app:example}

\begin{figure}[h]
\centering
\includegraphics[width=\textwidth]{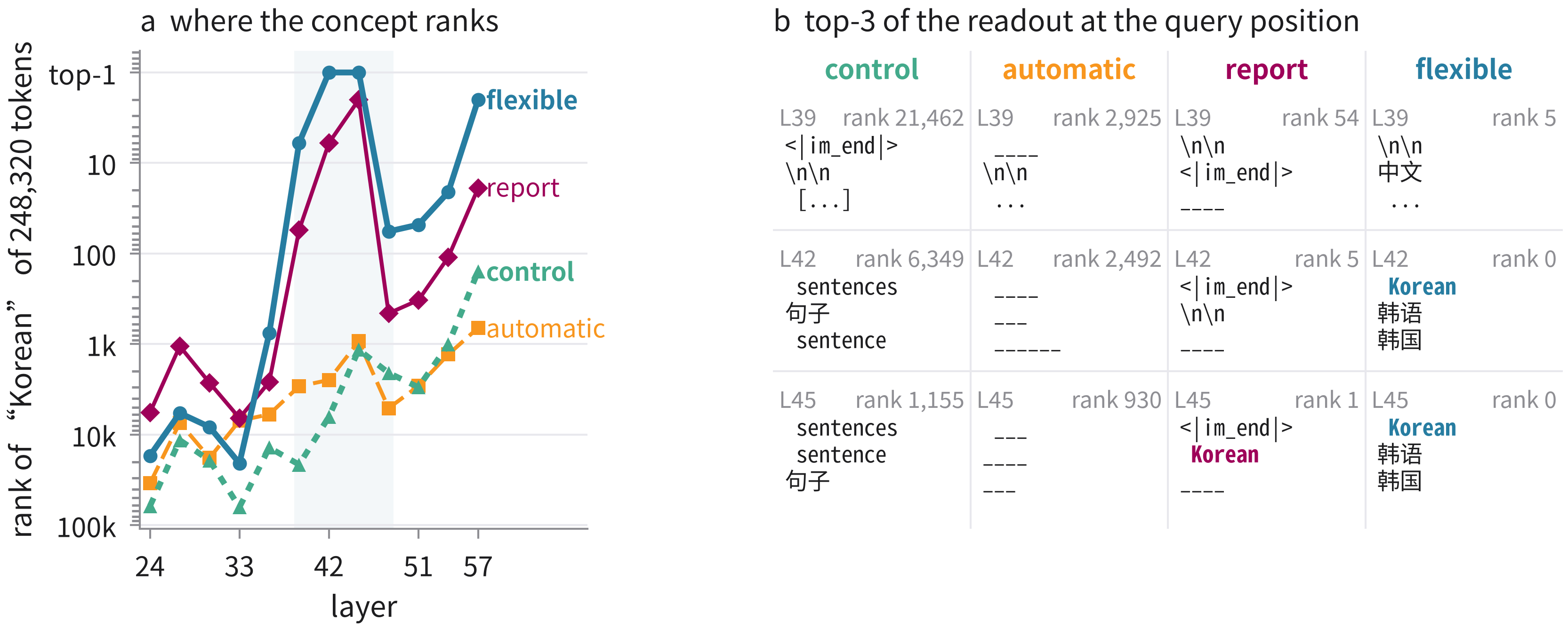}
\caption{One semantic instance, all four conditions, on Qwen3.6-27B. \textbf{(a)} The rank of the gold
concept (\emph{Korean}) among the vocabulary at the query position, by layer, on a log scale with top-1
at the top; shading marks the transport window. \textbf{(b)} The top three tokens of the readout at
the three layers where the arms diverge, with the gold token bold where it appears. The four arms ask,
respectively, whether the passage contains a question mark, for the next sentence, what language the
passage is written in, and which symbol corresponds to that language. Ranks in (b) count competing
tokens and so start at $0$, one below the $1$-indexed rank used everywhere else in the paper.
Token strings are shown
escaped, so \texttt{'\textbackslash n\textbackslash n'} is a literal double newline and a leading
space is visible.}
\label{fig:example}
\end{figure}

$\Rz$ is a percentile rank, which is precise and opaque. Figure~\ref{fig:example} shows what the
readout literally contains for one instance in all four arms.

First, the concept climbs to the top of the readout at L42--L45 in exactly the two arms that need it verbalizably (rank $0$ in flexible at both layers, and rank $5$ then rank $1$ in report), while at
those same layers the other two arms hold their own task's content instead. The control arm, asked
whether the passage contains a question mark, reads \texttt{' sentences'}, \texttt{' sentence'} and the
Chinese word for \emph{sentence}; the automatic arm, asked only to continue the passage, reads blanks.
Above L48 every arm's readout reverts to formatting tokens, which is why the band mean is the right
summary and a late readout is not. In the flexible arm at L42 the whole top of the readout is the same
concept in five surface forms across two scripts.

Second, and more usefully as a caution: at L45 the control arm's $\Rz$ is $0.9953$, which sounds like
strong representation. The gold token ranks \textbf{1155th} there, and nothing in the top of the readout
concerns language at all. A percentile close to $1$ over a $248{,}320$-token vocabulary is compatible
with a rank in the thousands, and the failure runs the other way too, since report reads
$\Rz = 1.0000$ at both rank $5$ and rank $1$. This is why \S\ref{sec:entry} reports the contrast and
never the absolute level, and why every mechanistic claim above is read under $\Lz$ as well.

\section{Artifacts and reproducibility}
\label{app:artifacts}

\paragraph{Configuration.}
The primary checkpoint is \texttt{Qwen/Qwen3.6-27B}: 64 layers, $d = 5120$, vocabulary $248{,}320$,
\texttt{full\_attention\_interval} $= 4$ so L3, L7, \dots, L63 carry standard attention with 24 heads of
width 256 and the other 48 layers a gated delta net with 48 value heads of width 128,
\texttt{attn\_output\_gate} $=$ true. The other four are \texttt{Qwen/Qwen3.5-9B},
\texttt{microsoft/phi-4}, \texttt{meta-llama/Llama-3.1-8B-Instruct} and
\texttt{google/gemma-4-31B-it}. Weights in bf16, all readout arithmetic in fp32;
\texttt{transformers} 5.14.1 with the \texttt{sdpa} attention implementation, except the
attention-pattern measurement of Appendix~\ref{app:transport}, which forces \texttt{eager} for one
module because \texttt{sdpa} discards the attention matrix. The flash-linear-attention fast path is
absent, so the gated delta layers run the reference torch implementation. Seed 0 everywhere except the
probe's five split seeds and the donor pairings, which are seeds $0$--$3$; instance selection is
deterministic (sorted by identifier, then truncated), so a rerun sees the same instances rather than a
fresh sample.

\paragraph{Exact revisions.}
So that a rerun can be pinned rather than approximated, the Hugging Face revisions actually loaded were
\texttt{Qwen/Qwen3.6-27B} at \texttt{6a9e13bd}, \texttt{Qwen/Qwen3.5-9B} at \texttt{c2022362},
\texttt{neuronpedia/jacobian-lens} at \texttt{a4114d77}, \texttt{haoranxu/FLORES-200} at
\texttt{8ecaf1bb} and \texttt{Salesforce/wikitext} at \texttt{b08601e0}; the tokenizer is each
checkpoint's own at the same revision. The published 27B lens file is $3{,}303{,}032{,}772$ bytes with
SHA-256 beginning \texttt{1718c8c5}, and the released five-arm language dataset is $1954$ records with
SHA-256 beginning \texttt{00623f40}. Every artifact the tables name is keyed to these.

\paragraph{Nothing is generated.}
No result in this paper decodes text. Every behavioural number is a forced choice: one forward pass,
then an \texttt{argmax} over the candidate token ids at the final position, so no decoding parameters
enter and no sampling nondeterminism with them. The only stochastic elements are the instance
split, the donor--target pairing and the bootstrap resampling, all seeded.

\paragraph{What each patch replaces, and where it is read.}
\texttt{resid.L$n$} substitutes the \emph{output} of block $n$: the residual stream after the block
has been added, before block $n+1$ reads it. \texttt{mlp.L$n$} substitutes the MLP submodule's output.
\texttt{attn.L$n$} is a forward \emph{pre}-hook on the attention output projection, so it replaces the
projection's \emph{input}: the per-head concatenation before mixing into the stream, which on the
primary checkpoint is the gated head output, \texttt{attn\_output} $\times$ $\sigma(\text{gate})$,
rather than the raw weighted value sum. \texttt{attn.L$n$.H$k$} replaces the $k$-th
\texttt{in\_features}$/n_{\text{heads}}$ slice of that input, computed per layer from the module rather
than from a configured \texttt{head\_dim}, because
$n_{\text{heads}} \times \text{head\_dim} \neq d_{\text{model}}$ on both branch types here.

\textbf{Where the patch goes and where the readout is taken are separate arguments}, and they have to
be: a multi-position patch otherwise drags the measurement off the retrieval cue, and the numbers stay plausible while they stop meaning anything. Every measurement here pins the
readout to the answer position independently of the patched span, which matters for the span-12 arms.
Mean ablation replaces a component's output with a \textbf{leave-one-out} mean
over up to 32 calibration records of the same arm; the record being ablated is removed from its own
baseline, as $(\sum x - x_i)/(n-1)$, so no item is ablated partly towards itself. Calibration records
always come from the \emph{same arm} of the same dataset at the same positions, never pooled across arms
and never from an external corpus, since a cross-arm mean would import the contrast being measured.

\paragraph{The lenses we fitted.}
Three of the five checkpoints run on lenses fitted here (Qwen3.5-9B, phi-4 and gemma-4-31B-it) and two on published artifacts. $J_\ell$ is a closed-form estimator rather than a trained model: a one-hot cotangent per output dimension is injected at each
valid target position and backpropagated, then averaged over valid source positions. Eight output
dimensions per backward pass, 512 backward passes per prompt, 120 prompts from
\texttt{Salesforce/wikitext} (\texttt{wikitext-103-raw-v1}, train, streamed, passages $\geq 400$
characters not beginning with \texttt{=}), \texttt{max\_seq\_len} 128, the first 16 positions skipped,
accumulated and saved in fp32. Passage distinctness is asserted at fit time rather than assumed. What we
cannot supply is the equivalent for the published 27B lens, which ships without a configuration file or
a convergence trace; that gap belongs to the artifact we consume, and it is why the replications fit
their own.

\paragraph{The audit.}
Every number here is written by a command-line entry point into a JSON or JSONL artifact, and every
figure carrying a measured quantity is generated from those artifacts rather than transcribed;
Figure~\ref{fig:overview} is the one exception, a hand-drawn schematic carrying no data. An audit
command shipped with the code machine-checks every numeric literal in this document against the
artifact tree; the few unmatched ones are software versions, artifact properties and derived
quantities. It is a floor
on transcription error and not a proof of correctness, so headline numbers are re-derived from their
own artifacts, and where a stored summary disagrees with its own observations we re-pool from the
observations.

We release the four-arm dataset and its five-arm extension, the generators, the lenses we fitted, and
the artifact path for every number, all MIT licensed. Because every prompt quotes a FLORES-200 passage
verbatim \citep{flores200}, records ship with the passage replaced by a language code and row offset;
one command rebuilds them byte-identically at the revision above, so we redistribute no share-alike
text and a rebuilt file is a FLORES derivative under \textbf{CC BY-SA 4.0}.

\end{document}